\pdfoutput=1
\documentclass[11pt]{article}

\usepackage{acl}

\usepackage{times}
\usepackage{latexsym}

\usepackage[T1]{fontenc}
\usepackage[utf8]{inputenc}

\usepackage{microtype}
\usepackage[normalem]{ulem}
\usepackage{times}
\usepackage{latexsym}
\usepackage{booktabs}
\usepackage[colorinlistoftodos]{todonotes}
\usepackage{enumitem}
\usepackage{xspace}
\usepackage{color, colortbl}
\usepackage{amssymb}% http://ctan.org/pkg/amssymb
\usepackage{pifont}% http://ctan.org/pkg/pifont
\usepackage{algorithm,algpseudocode}
\usepackage{amsmath}
\usepackage{amsfonts}
\usepackage{amssymb}
\usepackage{listings}
\usepackage{tikz}
\usepackage{makecell}
\newcommand{\secref}[1]{\S\ref{#1}}
\newcommand{\benchmark}{{\fontfamily{lmtt}\selectfont CLUES}\xspace}
\newcommand{\benchmarkreal}{{\fontfamily{lmtt}\selectfont \benchmark-Real}\xspace}
\newcommand{\benchmarksyn}{{\fontfamily{lmtt}\selectfont \benchmark-Synthetic}\xspace}
\newcommand{\model}{{\fontfamily{lmtt}\selectfont ExEnt}\xspace}
\newcommand{\quantmodel}{{\fontfamily{lmtt}\selectfont LaSQuE}\xspace}
\newcommand{\Thead}[1]{\textbf{\textsc{#1}}}
\newcommand{\probP}{\text{I\kern-0.15em P}}

\definecolor{LightCyan}{RGB}{172,204,186}

\makeatletter
\newcommand{\ostar}{\mathbin{\mathpalette\make@circled\star}}
\newcommand{\make@circled}[2]{%
  \ooalign{$\m@th#1\smallbigcirc{#1}$\cr\hidewidth$\m@th#1#2$\hidewidth\cr}%
}
\newcommand{\smallbigcirc}[1]{%
  \vcenter{\hbox{\scalebox{0.77778}{$\m@th#1\bigcirc$}}}%
}
\makeatother

\title{\quantmodel: Improved Zero-Shot Classification from Explanations Through Quantifier Modeling and Curriculum Learning}
\author{Sayan Ghosh\thanks{\hspace{0.5em}Equal contribution} \and Rakesh R Menon\footnotemark[1]  \and Shashank Srivastava\\
   UNC Chapel Hill \\ 
  \texttt{\{sayghosh, rrmenon, ssrivastava\}@cs.unc.edu}}

\begin{document}
\maketitle

\begin{abstract}
A hallmark of human intelligence is the ability to learn new concepts purely from language. 
Several recent approaches have explored training machine learning models via natural language supervision. However, these approaches fall short in leveraging linguistic quantifiers (such as `always' or `rarely')
and mimicking humans in compositionally learning complex tasks. Here, we present \quantmodel, a method that can learn zero-shot classifiers from language explanations by using three new strategies - (1) modeling the semantics of linguistic quantifiers in explanations (including exploiting ordinal strength relationships, such as `always' > `likely'), (2) aggregating information from multiple explanations using an attention-based mechanism, and (3) model training via curriculum learning. With these strategies, \quantmodel outperforms prior work, showing an absolute gain of up to 7\%
% \sout{on the recently proposed \benchmark benchmark} 
in generalizing to unseen real-world classification tasks.\footnote{Work in progress.}

\end{abstract}
\section{Introduction}

%% Para 1: What is the task? What is learning from language. How is it usually done?

% What is the task?

% \todo{I like this version of the introduction much better!}
% \ssr{Will read through and comment on the intro later}
Learning from language (also `conversational machine learning') is a new paradigm of machine learning where machines are taught tasks through natural language supervision in the form of explanations and instructions \cite{andreas-etal-2018-learning, arabshahiconversational, weller-etal-2020-learning, efrat2020turking}.
Language explanations of concepts have been explored for training classification models in few-shot and zero-shot settings \cite{mei2022falcon, srivastava2017joint,srivastava-etal-2018-zero, hancock-etal-2018-training, desc_based_classification, obeidat-etal-2019-description, hanjie2022semantic}.
% and zero-shot settings \cite{srivastava-etal-2018-zero}. 
% \textcolor{blue}{
%Taking it one step further, \citet{anon_benchmark} introduced zero-shot models that can perform classification tasks purely from language explanations without access to any relevant unlabeled data.
% }
% More recently, \citet{anon_benchmark} introduced a benchmark, \benchmark, for learning zero-shot classifiers from explanations in a cross-task generalization setup and show that their entailment-based model, \model, can perform novel classification tasks purely from language explanations. \ssriv{Prev sentence positions your work as being derivative of CLUES without adding much -- there's no need to refer to CLUES at this point. }

\begin{figure}
    \centering
    \includegraphics[scale=0.35]{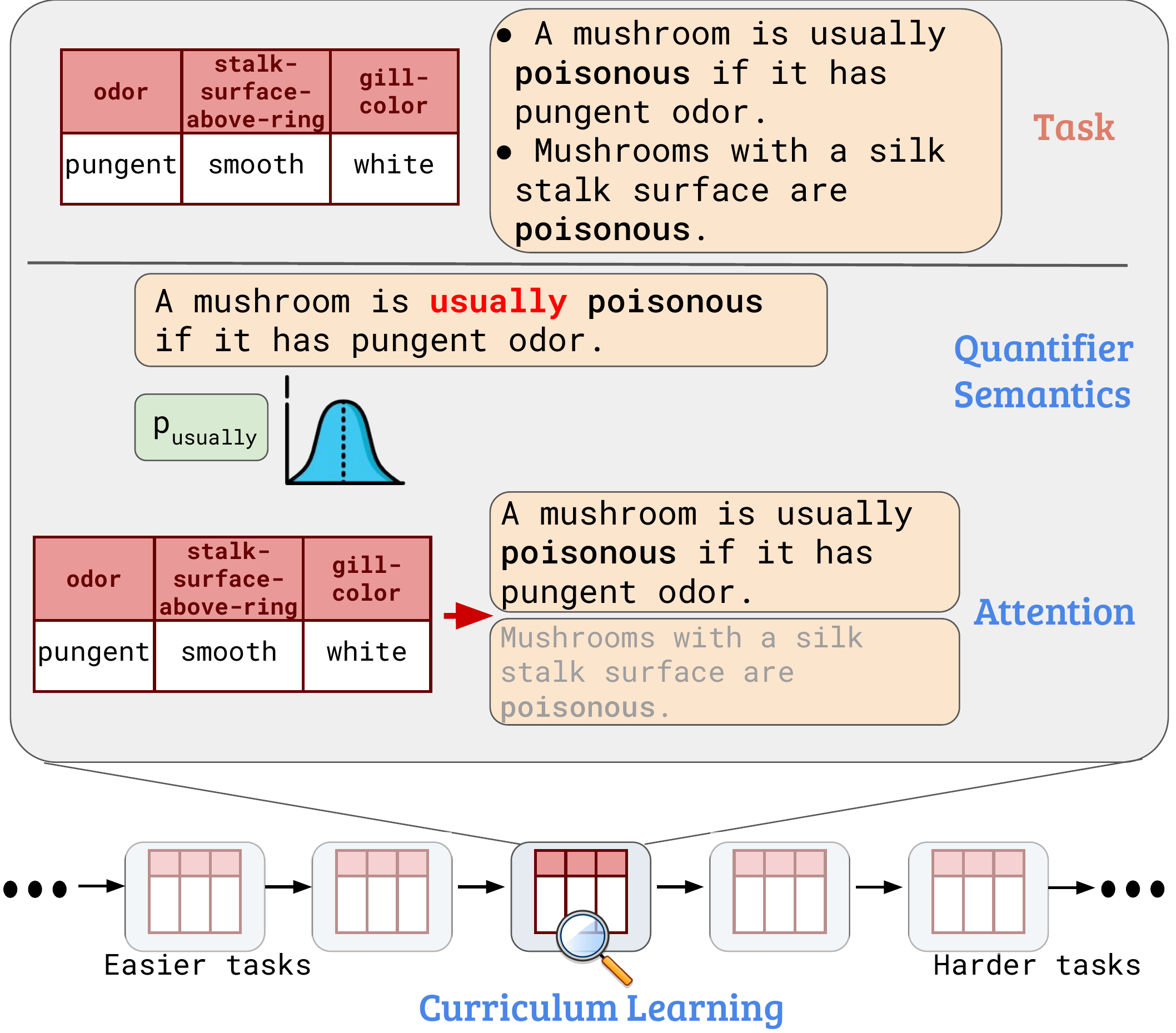}
    \caption{We present techniques to learn zero-shot classifiers from natural language explanations. We investigate (1) learning Quantifier Semantics to fully leverage supervision from individual explanations, (2) Attention-based mechanisms to identify the most salient explanations for learning a task, and (3) Curriculum Learning to learn more complex tasks progressively. 
    % \ssc{Issue with the figure, the main panel is more or less repeated }
    % (ii) \sout{the order of things in the figure should roughly mirror the order in which we discuss them in the paper. Currently, curriculum learning is sitting at the top, while we talk about it near the end of the paper}
    % \ssr{Curriculum Learning and Attention are noun phrases, whereas Modeling Quantifier Semantics is a verb phrase. Replace with Quantifier Semantics or Quantifier Semantics Learning?. Also, there is a robot is two panels, but its unclear what its role is (without a human in another panel) -- either have both or neither }
    \vspace{-0.65cm}
    }
    \label{fig:intro_figure}
\end{figure}

%% Para 2: What are the shortcomings of the current approaches to learning from language?

%While recent approaches like \model show promise in learning from explanations, they still 
However, current approaches fall short in fully leveraging supervision available in language explanations and using learning strategies that humans routinely employ in learning new tasks. % modeling the intricacies of natural language or in mimicking human behaviour in learning a suite a tasks with varying difficulty.
First, most approaches, such as LNL \cite{srivastava2017joint}, and BabbleLabble \cite{hancock-etal-2018-training}, do not model supervision within free-form language explanations in the form of quantifiers. 
% \ssriv{Again, why pit this directly again ExEnt. You can include (multiple) general citations here}. 
% \rrm{Say more directly fails to capture quantifiers within free-form language...}
% \rrmenon{break}
% \textcolor{blue}{
Quantifiers are linguistic elements that can dictate the vagueness and perceived confidence of relations expressed in a statement~\cite{Solt2009THESO, quantifier_focus}. For example, with statements such as \textit{`some poisonous mushrooms are red in color'}, we can infer that a red mushroom is not always poisonous because of the quantifier \textit{`some'}.
Moreover, quantifiers are a ubiquitous part of natural language and universal across languages.
% }
% Moreover, quantifiers are a ubiquitous part of natural language, which is evidenced by the presence of quantifiers in nearly half of the explanations of \benchmark benchmark \ssriv{Last sentence -- remove reference to CLUES and just say they are universal across languages}\rrmenon{find another set of citations}. 
Second, prior approaches do not reason about differences in salience and utility of multiple explanations in learning a new task, weighing them equally in the absence of labeled data. This is sub-optimal since certain explanations can be naturally harder to incorporate or have inherently less value in learning a concept\footnote{e.g., overly complex, or highly subjective explanations}. Thirdly, when learning a set of tasks, humans often learn `simpler' concepts first and gradually build towards `harder' concepts~\cite{newport1990maturational}. While curriculum learning~\cite{bengio2009curriculum} has been shown to be fruitful in numerous machine learning tasks~\cite{platanios2019competence,tay2019simple}, %,narvekar2017autonomous}, 
it is yet to be explored in the context of learning from explanations. The deteriorating generalization of classifiers with the increasing complexity of explanations in prior work~\cite{anon_benchmark} further motivates curriculum learning for learning from explanations. 
% \ssc{While this statement is  relevant, the intro seems too heavily focused on our own work and not enough on other work}

%% Para 3: What are the improved techniques that we will look to add in the paper?
% \rrmenon{Maybe we can list these as points to make it easier to read.}
% To address the first shortcoming, our approach \quantmodel explicitly models quantifier semantics. 
To address the first shortcoming, our approach \quantmodel \footnote{Short for \textbf{L}e\textbf{a}rning \textbf{S}trategies For \textbf{Qu}antified \textbf{E}xplanations} explicitly models quantifier semantics and learns them directly from labeled classification data. However, directly learning from labeled data can lead to quantifier semantics that is inconsistent with human perceptions of their numerical estimates. Hence, we provide weak supervision in the form of ordinal relations describing the relative strengths of quantifiers  (e.g., `always' > `likely') to supplement the learning of quantifier semantics that comply with human judgments.
% We explore learning quantifier semantics directly from labeled classification data and weak supervision in the form of ordinal relations describing the relative strengths of quantifiers  (e.g. `always' > `likely'). \rrmenon{add emphasis for novelty}
Second, we design an attention-based mechanism to model the relative importance of multiple explanations in classifying an example. % to identify salient explanations that are helpful towards classifying an input. 
We also qualitatively analyze the attention weights to identify characteristics of explanations found most helpful. % considered helpful by \quantmodel. % to be more
% \quantmodel outperforms prior work showing an absolute improvement of upto 5.2\% in generalization accuracy on classification tasks unseen during training. \rrmenon{Should come after curriculum because quexent is bad till here.}
Finally, we consider different axes of explanation complexity and empirically evaluate the utility of curriculum learning on three different curricula. 
% Overall, we find curriculum learning to be effective in improving the generalization performance of classifiers by 17\% 
% (absolute)\ssriv{Why does this have 2 decimal digits reported?}
% for tasks with complex explanations, especially when explanations contain negations. \rrmenon{mention this is for synthetic or rephrase this last part.}

\noindent
% \rrmenon{Introducing CLUES this way seems odd. We mention results are boosted without mentioning ``on what dataset/benchmark" in the previous paragraph. Some restructuring is needed here as well.}
As our test bed, we use the recently proposed \benchmark benchmark~\cite{anon_benchmark} for learning classification tasks from language explanations. Our work focuses on learning classifiers from language explanations where the explanations provide the logic to perform the classification. (e.g., the explanation `pungent mushrooms are toxic', provides the logic that mushrooms with a pungent odor should be classified as toxic). 
\benchmark is the largest available benchmark that contains explanations conformant with this perspective. It differs from some other
%Hence, \benchmark by virtue of its design is well suited for our study\footnote{Also the largest available benchmark under the said notion of explanations.}. As highlighted in \citet{anon_benchmark}, \benchmark is different from other existing 
benchmarks~\cite{Mishra2021CrossTaskGV, sanh2021multitask}, where the language component provides the \emph{description of the task} instead (such as, `classify the mushrooms as toxic or poisonous'), which can be used to train/prompt a model. 
% \textcolor{blue}{
On \benchmark, \quantmodel achieves an improvement of 17\% and 7\%, respectively, on the synthetic and real-world benchmarks over baselines.
% }
% (absolute)\ssriv{Why does this have 2 decimal digits reported?}
% for tasks with complex explanations, especially when explanations contain negations. 

%% Para 4: Summarize contributions
The rest of this paper is structured as follows: we provide a description of the preliminaries in \secref{sec:background}.
% including details of \model 
In \secref{sec:quexent} we describe \quantmodel and our learning strategies 
along with supporting empirical performance. \secref{sec:real_world_results} discusses performance of \quantmodel on real world classification tasks.
Our contributions are:
\begin{itemize}[noitemsep, topsep=0pt, leftmargin=*]
    \item We introduce \quantmodel, which models semantics of linguistic quantifiers, and uses %present in natural language explanations for the task of learning classifiers from explanations. 
    %We use 
    an attention-based mechanism %in \quantmodel 
    to identify salient explanations for learning classifiers from language.
    \quantmodel significantly outperforms previous methods in generalizing to unseen classification tasks. % of the \benchmark.
    \item We %present three different curricula over synthetic tasks of \benchmark benchmark 
    empirically demonstrate the utility of curriculum learning in training classifiers from language by experimenting with three curricula.
    % \rrmenon{Should we mention we advance the best performance on clues? (in some way; not specifically as a sole point)}
    % \item We test the limits the pre-training on synthetic classification tasks before fine-tuning on real-world classification tasks. 
\end{itemize}

%% Why CLUES?
% \sayan{no need for this separate para. include this in the setup}
% \paragraph{Why \benchmark?} In this work, we base our analyses on the recently proposed benchmark, \benchmark \cite{anon_benchmark}. This is motivated by the fact that \benchmark is the largest available benchmark for classification tasks associated with natural language explanations. Consequently, the benchmark allows for studying models in a cross-task generalization setting \cite{Mishra2021CrossTaskGV} \ssr{unclear why you cite this specific paper here} and designing classifiers that can solve novel tasks \textit{purely} from language. Further, \benchmarksyn, the synthetic benchmark within \benchmark, defines task sets with explicit demarcations of difficulty levels based on explanation complexity. This provides us with the unique opportunity to ablate and identify a model's capability in solving specific complexities.
% \rrm{Is there a particular other dataset that could be possibly be used? If so, maybe you should mention that dataset and why you are not using it? If not, why have this section?} 
\section{Related Work}

% \paragraph{Learning concepts from auxiliary information and explanations.} The use of `side-information' to guide models during training has been previously explored in \citet{mann2010generalized, ganchev2010posterior}. However, these approaches require users to specify auxiliary information/constraints using mathematical expressions \ssr{I don't think GE and PR are too relevant here. Also, the wording seems very similar to my previous papers}. Recently, \citet{srivastava2017joint, hancock-etal-2018-training} apply semantic processing to convert user-provided natural language statements to constraints for learning in limited labeled data settings. In contrast to these works \ssr{I don't see what the contrast is. At the most, you can say that you additionally model quantifier semantics. The contrast might be in terms of the methods and representations -- semantic parsing vs distributed representations -- but that's not what you say. Also, the CLUES paper already does this.}, we explore the use of quantifier information in enabling learning from language models generalize to tasks with no labeled examples.\ssr{This subsection needs more work or should be removed}

\paragraph{Natural Language Quantification.} 
Previous work has studied the role of quantifiers in natural language from multiple perspectives, such as formal logic \cite{barwise1981generalized}, linguistics \cite{lobner1986quantification, bach2013quantification}, cognitive psychology \cite{kurtzman1993resolution}, and natural language processing to guide statistical models \cite{srivastava-etal-2018-zero}. In the above mentioned works, quantifiers have been typically modelled in either set-theoretic terms \cite{barwise1981generalized} or by representing them probabilistically \cite{moxey1993prior, yildirim2013linguistic, srivastava-etal-2018-zero}. 
Our work is closely related to \citet{srivastava-etal-2018-zero}, who also model the effects of quantifiers in modifying the belief of a classifier. However, we differ from \citet{srivastava-etal-2018-zero} as we learn the beliefs associated with quantifiers during training as opposed to defining them apriori with fixed values.
More recently, \citet{cui-etal-2022-generalized-quantifiers} discusses the challenges in understanding quantifiers, specifically in the context of NLI, and contributes a focused test dataset to benchmark NLI models on their ability to understand quantifiers. While both \citet{cui-etal-2022-generalized-quantifiers} and our work broadly highlight the need to model quantifiers to advance language understanding, we differ in the nature of downstream tasks studied (diverse classification tasks in our work vs NLI in \citet{cui-etal-2022-generalized-quantifiers}). 
% Our work is closely related to \citet{srivastava-etal-2018-zero} where the effect of quantifiers in modifying the belief of a classifier is also modeled probabilistically. However, we differ from \citet{srivastava-etal-2018-zero} as we learn the probability values associated with quantifiers as opposed to defining them as hyperparameters beforehand. 
Our approach (\quantmodel) contains a dedicated module that enables us to \emph{learn} quantifier semantics, which apply to a wide range of tasks spanning multiple domains. 
% \rrm{I think we need to stress on the fact that our model, while requiring some training data, is more applicable across a range of domains while LNQ is not because it relied on domain-specific semantic parsers.}

% \rrm{Add a para on curriculum learning related works?} \ssr{Yes, should definitely do that.}

\paragraph{Curriculum Learning.} 
Curriculum learning \cite{bengio2009curriculum} is a technique to learn complex tasks through a graded exposure of examples ranging from easy-to-hard difficulty. Recent works in machine learning \cite{jiang2018mentornet, guo2018curriculumnet, hacohen2019power} have successfully demonstrated the utility of curriculum learning in learning image classification tasks. More recently, \citet{xu-etal-2020-curriculum} also demonstrated the effectiveness of curriculum learning for a set of natural language understanding tasks drawn from the GLUE benchmark \cite{wang-etal-2018-glue}. However, prior works build a curriculum of easy-to-hard examples to improve model performance on individual tasks. Rather than examples, we build a curriculum of easy-to-hard tasks in our work, similar to \citet{mao2018neuro}. In contrast to \citet{mao2018neuro} though, we focus on learning structured data classification tasks from language explanations as opposed to visual question answering tasks.
\section{Preliminaries}
\label{sec:background}
% \ssr{The name of this section needs to change. This looks like related work again. Maybe preliminaries?}
\subsection{Setup}
\label{sec:benchmark_setup}
We employ a cross-task generalization setup \cite{Mishra2021CrossTaskGV, sanh2021multitask}, and train models using multi-task training over a set of tasks $\mathcal{T}_{seen}$ and evaluate for zero-shot generalization on a new task, $t \in \mathcal{T}_{novel}$ ($\mathcal{T}_{novel} \cap \mathcal{T}_{seen} = \phi)$. The evaluation metric is the zero-shot classification accuracy on novel classification tasks. For experiments, we use the recently proposed \benchmark benchmark~\cite{anon_benchmark}. The benchmark is composed of synthetic and real-world classification datasets. The real-world classification tasks were created using resources from UCI Machine Learning repository, Kaggle, and Wikipedia tables. The explanations for real-world tasks were crowdsourced. The synthetic tasks were created programmatically to study the performance of models under different levels of task complexities. The 48 different complexities in \benchmarksyn arise from the: (a) presence of negations in clauses and/or labels, (b) structure of explanations (conjunctions/disjunctions/nested), (c) presence of quantifiers in explanations, and (d) binary vs multiclass classification task. The explanations for \benchmarksyn are generated programmatically using templates.

In \benchmark, inputs are structured, consisting of attribute name-attribute value pairs (see Figure \ref{fig:intro_figure} for example). 
We use the `Features-as-Text' or `FaT' representation to encode the examples following \citet{anon_benchmark}.
Additional details about \benchmark can be found in Appendix \ref{sec:appendix}. We use the following two baselines in our experiments: (1) RoBERTa w/o Exp (does not use explanations) \cite{liu2019roberta} and (2) \model (uses explanations). Next, we describe \model in more detail. % in \secref{sec:old_model}.

\subsection{\model}
\label{sec:old_model}
\citet{anon_benchmark} identified that a simple concatenation of explanations with the input was insufficient to enable pre-trained language models to generalize to novel tasks. Therefore, they introduce \model, a model which uses Natural Language Inference (NLI) as an intermediate step, and broadly operates through the following steps: %. The operations in \model can be broadly grouped into three steps: 
% \ssr{Remove bold font since what you are describing is existing work, also 1/2/3 is better since than abc, esp. since you say `three steps' } 
(1) NLI step: obtain scores from an entailment prediction model (RoBERTa+MNLI-finetuned) for the alignment between the input and each explanation available for a task; (2) Entailment $\rightarrow$ Classification scores conversion: convert the entailment scores for each input-explanation pair into classification scores based on the nature of the explanation; and (3) Aggregation: average the classification scores from each input-explanation pair to obtain an aggregate score for classification.
% Finally, the scores are converted to probabilities using a softmax operation, and the model is trained using the cross-entropy loss. 
% The aggregated scores are then converted to probabilities using softmax, and the model is trained using cross-entropy loss. 
Convert aggregate scores to probabilities using softmax and train the model end-to-end using the cross-entropy loss. 
For more details, we refer the reader to \citet{anon_benchmark}.

\begin{figure*}[!t]
    \centering
    \includegraphics[scale=0.6]{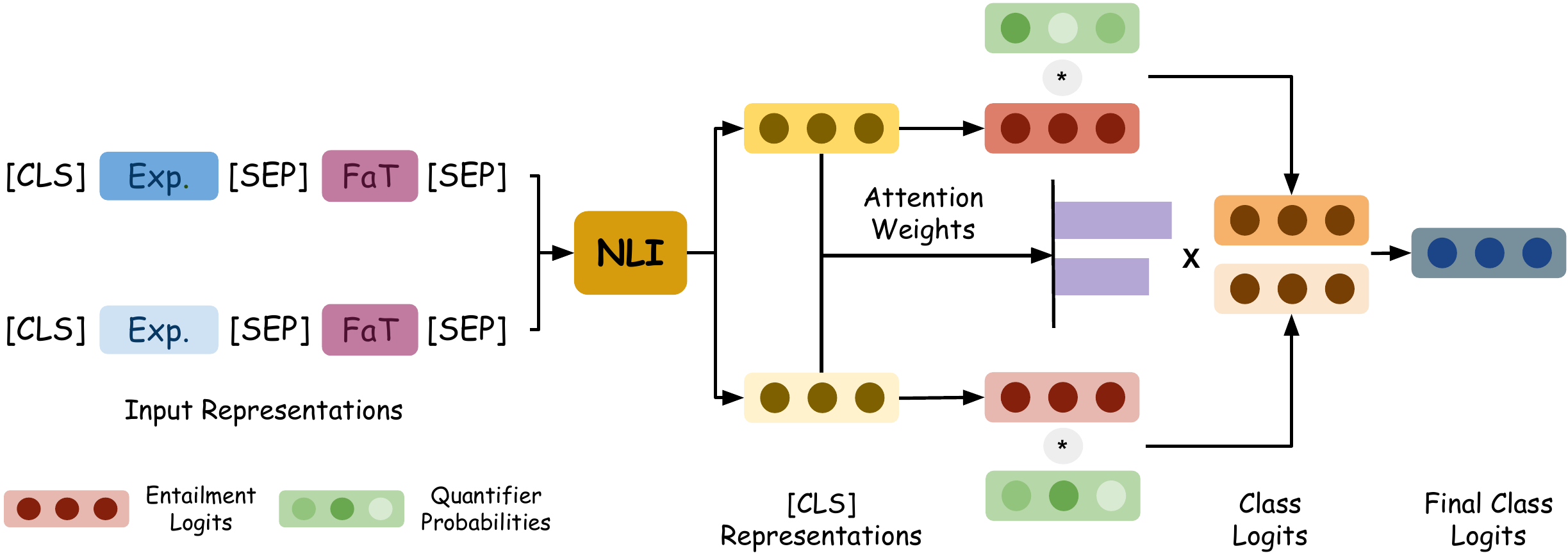}
    \caption{\quantmodel models quantifier semantics and uses attention over multiple explanations to aggregate class logits. As shown in the figure, our approach allows us to re-weight the logits from the NLI step, thus strengthening/weakening the contribution of an explanation towards assigning the label (mentioned in the explanation) to the input. $\ostar$ denotes the operations described in \secref{sec:new_model} for assignment of class logits using the outputs from the NLI step. Curriculum learning (not shown in the figure) entails training \quantmodel progressively on easy-to-hard tasks.}
    % \sayan{mention attention in figure; curriculum in caption}
    % \ssr{This figure can be made more compact. Currently, too much white space and vertical spacing. e.g., the two vectors at the bottom could be made smaller and brought up/ simply labeled within the main figure. The other concern is this looking too much like the CLUES paper: maybe make some minor changes in the color patterns? Also, maybe remove boundary lines for all the shapes?}}
    
    \label{fig:proposed_model}
\end{figure*}

\section{\quantmodel}
\label{sec:quexent}
In this section, we present our method, \quantmodel, and provide detailed descriptions and empirical support for the different learning strategies that are part of \quantmodel -- (1) modeling quantifier semantics, (2) using attention for aggregation across explanations, and (3) curriculum learning.

\subsection{Modeling quantifier semantics}
\label{sec:new_model}
Quantifiers are a ubiquitous part of natural language and can help express varying strengths of relations in a statement. Prior work in cognitive science \cite{Chopra2019TheFC, steinert2021quantifiers} and machine learning \cite{srivastava-etal-2018-zero, anon_benchmark} shows that people tend to use quantifiers often in learning or teaching tasks.  Hence, modeling quantifiers is important for building systems that can mimic humans in efficiently learning from natural language. However, past work on computational modeling of quantifiers is sparse. To the best of our knowledge, no prior work has explored learning quantifier semantics in a data-driven way. In this work, we devise methods to explicitly model the differential semantics of quantifiers present in explanations to guide classifier training.
Figure~\ref{fig:proposed_model} shows architecture of our model, \quantmodel.
% We train and evaluate our method \quantmodel on the \benchmark benchmark\ssr{You have already discussed using CLUES before. Also, you were discussing the model, and the next sentence again talks about the model as well. The previous sentence is both (1) superfluous and (2) out-of-place here}.
% \vspace{-0.1cm}

To formalize our approach to modeling quantifier semantics, consider a task $t$ with the set of class labels $L$ and set of explanations $E$. Given the Feature-as-Text (FaT) representation of a structured data example $x \in t$ and an explanation $e_j \in E$, our model takes FaT($x$) and $e_j$ as input and passes it through a pretrained RoBERTa+MNLI model, similar to previous work~\cite{anon_benchmark}. For each example-explanation pair, the NLI model outputs entailment, neutral, and contradiction scores (denoted as $s^j_e$, $s^j_n$, and $s^j_c$ respectively). In the next step, we incorporate quantifier semantics to assign logits 
% \rrm{Do these logits serve as weigths? It might be a bit more clear if you use weights?}
to the set of class labels, $L$, using the outputs of the NLI model. In this work, we model the semantics of a quantifier by a probability value signifying the strength of the quantifier, i.e., the confidence of the quantifier in conveying the beliefs expressed in the explanation.
The assignment of class logits is done as follows. If:
\begin{itemize}[noitemsep, topsep=0pt, leftmargin=*]
    \item \textbf{Explanation $e_j$ mentions a label $l_{exp}$}: 
    An illustrative example is `If head equal to 1, then dax', where `dax' is the label mentioned ($l_{exp}$).
    Let $p_{quant}$ denote the probability of the quantifier mentioned in the explanation\footnote{We assume that explanations contain a single quantifier. This assumption also holds true with the explanations found in \benchmark.} and $\mathbb{P}(l)$ denote the probability of any label $l \in L$. Then,
    \begin{multline}
    \small
        \log(\mathbb{P}(l_{exp})) \propto p_{quant} \times  s^j_e  \\ + (1 - p_{quant}) \times  s^j_c + s^j_n/|L|
    \end{multline}
    \begin{multline}
    \small
        \forall~l \in L \setminus \{l_{exp}\}, \\ \log(\mathbb{P}(l)) \propto p_{quant} \times  s^j_c \\  + (1 - p_{quant}) \times  s^j_e + s^j_n/|L|
    \end{multline}
    
    \uline{Note}: If quantifiers are absent in the explanations, we assume $p_{quant}$ is 1.
    
    \item \textbf{Explanation $e_j$ mentions negation of a label `$l_{exp}$' (NOT $l_{exp}$)}: 
    An illustrative example is `If head equal to 1, then not dax", where `dax' is the label mentioned ($l_{exp}$).
    The roles of $s^j_c$ and $s^j_e$ as described in the previous equations are reversed.
\end{itemize}
% We refer to the logit assigned to a label $l \in L$ the explanation $e_j$ as $z_j^l$ henceforth. 
% \ssr{Remove this sentence. You are not using this name anywhere close. You are redefining it in 4.2 anyhow}

Following this step, we average the class logits from each example-explanation pair to aggregate the decisions. Finally, we apply a softmax over the resulting class scores to obtain a distribution over class labels and train the model to minimize the cross-entropy loss, $\mathcal{L}_{CE}$.

\begin{figure*}[!h]
    \centering
    \includegraphics[scale=0.48]{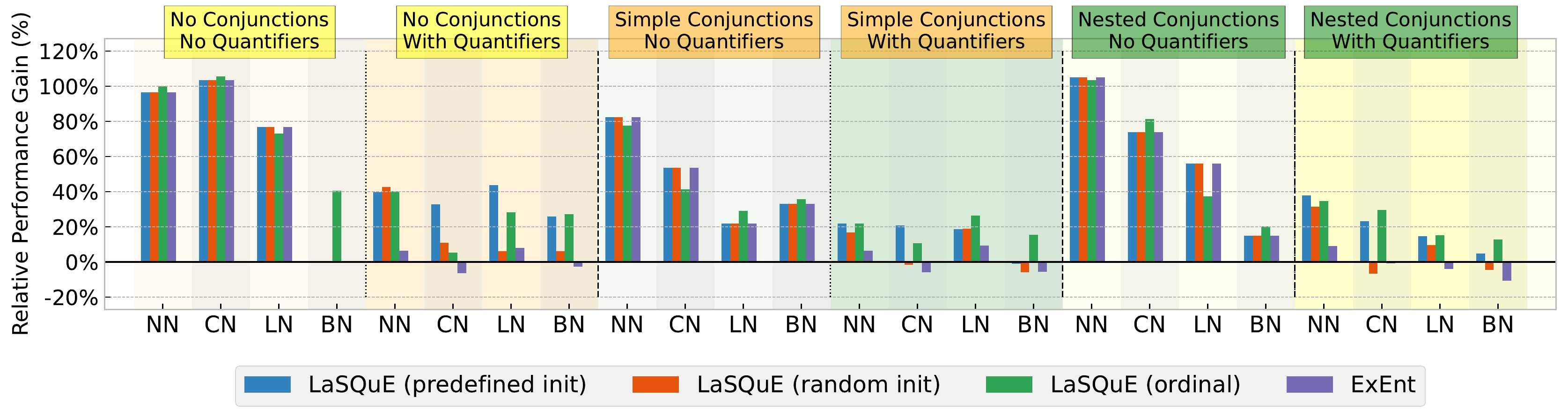}
    \caption{Relative performance gain of \quantmodel and \model with respect to the baseline RoBERTa w/o Exp. model. `NN', `CN', `LN', and `BN' stand for `no negations', `clause negations', `label negations', and `both negations' respectively denoting the variations of negations appearing in the explanations of \benchmarksyn.
    We see that \quantmodel outperforms \model~\cite{anon_benchmark} on all tasks having explanations with quantifiers. 
    Within each panel, the complexity of tasks increases from left to right due to negations. Across panels, the complexities also increase from left to right due to a combination of change in the presence of quantifiers and explanation structure (using conjunctions/disjunctions). 
    % conjunctions and quantifiers.
    % Each panel specifies the complexity of explanation based on its structure and presence of quantifiers, and the complexity of panels increase from left to right. Within each panel, the complexity increases from left to right due to presence of negations in explanations. 
    % The generalization performance decreases with increase in explanation complexity.
    % \ssr{you mean complexity increases within each panel from l to r, or b/w the six panels from left to right? Or both? Also clarify the same in the text}) due to presence of conjunctions, disjunctions and/or negations.
    % \ssr{In the text, you have been using synthetic tasks, this seems to be the first use of synthetic dataset. Is a synthetic dataset same as a synthetic task, is it the same as a task type?} that contain explanations with quantifiers. 
    % \ssr{In the text, you should also comment on that Quexent alwqays outperforms Roberta w/o explanations, while ExEnt is actually weaker in some cases.}
    % \ssr{Text is too small to read. You can also consider using abbreviations in the figure (NN, CN, LN, BN) and explaning them in the caption. }
    }
    % \vspace{-0.3cm}
    \label{fig:exent_q_all_synthetic}
\end{figure*}

\paragraph{Approaches to learn quantifier semantics.}
\label{sec:approaches_quantifier_semantics}
We experiment with the following approaches to learn the probability values quantifiers:
\begin{itemize}[noitemsep, topsep=0pt, leftmargin=*]
    \item \uline{Finetuning pre-defined probability values}: 
    We initialize the quantifier probability values ($p_{quant}$) with pre-defined values and fine-tune them while training \quantmodel. These initial estimates can be specified from domain knowledge or by an expert. In this work, we adopt the quantifier values from \citet{srivastava-etal-2018-zero}.\footnote{Full list of quantifiers used can be found in the Appendix.}
    We refer to the model learned using this approach as \quantmodel (predefined init).
    
    \item \uline{Learning probability values for the quantifiers from scratch}:
    We start from random initialization and then learn the probability values of each quantifier while training \quantmodel. 
    We refer to the model learned using this approach as \quantmodel (random init).
    % \rrm{Is the only difference between this and above the initalization?}
    
    \item \uline{Ordinal ranking as weak supervision}: 
    % \ssr{This section is also vaguely written. Try to rewrite it by including equations.}
    We explore another form of supervision by specifying ordinal relationships between pairs of quantifiers based on their relative strengths. To define ordinal relationships, we re-purpose the quantifier probability values in \citet{srivastava-etal-2018-zero}. For example, quantifiers such as `likely' and `often' associated with the values 0.7 and 0.5 respectively are defined by the relationship, `likely' > `often'.
    We leverage the ordinal relations to guide the learning of quantifier semantics through a ranking loss, following \citet{pavlakos2018ordinal}. Given a pair of quantifiers $q_i$ and $q_j$ $(i \neq j)$, the ranking loss is defined as:
    \begin{align*}
    % \small
        \mathcal{L}_{i,j}= \left\{
                            \begin{array}{@{}l@{}}
                            \log(1 + \exp(p_{q_i} - p_{q_j})), \hspace{1em} \mathbf{p^*_{q_i}} > \mathbf{p^*_{q_j}}\\
                            \vspace{0.1em}
                           (p_{q_i} - p_{q_j})^2, \hspace{6em} \mathbf{p^*_{q_i}} = \mathbf{p^*_{q_j}}
                           \end{array}
                           \right.
    \end{align*}
    where, $\mathbf{p^*_q}$ refers to the subjective probability value of a quantifier, $q$, in \citet{srivastava-etal-2018-zero}. Further, we define
    \begin{equation}
    % \small
    % \vspace{-0.15cm}
        \mathcal{L}_{rank} = \sum_{(q_i, q_j) \in Q} \mathcal{L}_{i,j}
    % \vspace{-0.1cm}
    \end{equation}
    where, $Q$ denotes the full set of quantifiers present in the explanations of \benchmark (\secref{sec:tab_quantifiers}). 
    The final loss is a weighted sum of classification loss ($\mathcal{L}_{CE}$) and ranking loss ($\mathcal{L}_{rank}$).
    \begin{equation}
    % \small
        \mathcal{L}_{total} = \mathcal{L}_{CE} + \lambda \mathcal{L}_{rank}
    \end{equation}
    where, $\lambda$ denotes the weight of ranking loss. We use $\lambda=10$ in this work, chosen using validation performance.
    We refer to the model learned using this approach as \quantmodel (ordinal). \\
    % \ssr{You can simply use $\lambda$ unless there are other $\lambda$s in the paper. Also, describe the value used or how it was chosen (e.g. from dev set)}
    
\end{itemize}

\begin{figure*}[!h]
    \centering
    \includegraphics[scale=0.45]{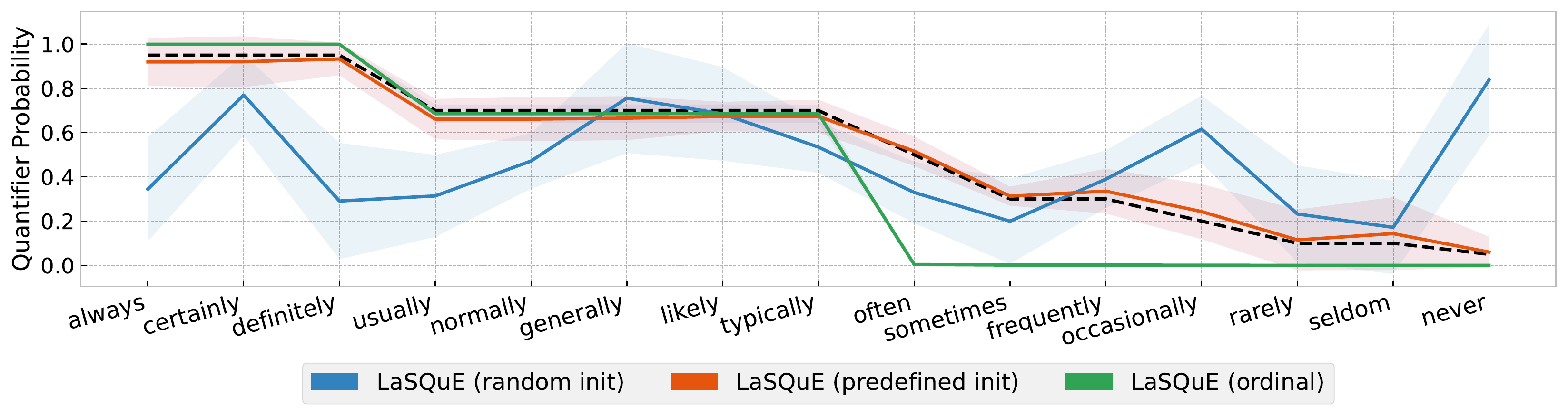}
    \caption{Quantifier probability values learned by the different approaches mentioned in \secref{sec:approaches_quantifier_semantics}.
    The solid line and the shaded region denote the mean and standard deviation respectively of the learned probabilities for a given approach across 48 synthetic task complexities in \benchmarksyn.
    The dotted-line denotes (1) probability values used by \citet{anon_benchmark} to create synthetic tasks of \benchmarksyn and (2) the quantifier probability initialization values for \quantmodel (predefined init).
    % \rrm{Can we stop this caption here? And have the next line in the body?}
    % The learned probabilities from \quantmodel (predefined init) after fine-tuning are close to the initial values. \quantmodel (ordinal) learns three clusters of quantifier probabilities, our intuition of high-strength, low-strength, and intermediate-strength quantifiers, but makes little difference between quantifiers within a cluster.
    % identifies differences between extreme and non-extreme quantifiers but is unable to learn subtle differences between non-extreme quantifiers 
    % \ssr{The last sentence (`unable to learn subtle differences]) is problematic -- it seems to suggest that the right thing to do would be to differentiate b/w all quantifiers, or in other words that there is a `gold-annotation' for quantifier semantics. Those numbers in the Srivastava18, Menon22 papers were just subjectively chosen. Better to phrase this as an observation, something like `We find that the ordinal method learns three clusters of quantifier probability values, our intuition of high-strength, low-strength and intermediate-strength quantifiers, but makes little differences b/w quantifiers within a cluster'} 
    % \ssr{Also, is the reason for there being no shaded region for the ordinal line?}\sayan{std devs for this are close to 0}. 
    }
    % \vspace{-0.3cm}
    \label{fig:quantifier_weights}
\end{figure*}

\noindent \textbf{Performance on \benchmarksyn: }
To evaluate the effectiveness of natural language quantification in learning classifiers from language explanations, 
we experiment on a collection of 100 tasks for each of the 48 different complexities from \benchmarksyn (see \secref{sec:benchmark_setup} for descriptions of task variations). 
We train a classifier for each of the different complexities and evaluate its generalization to novel tasks of same complexity.

Figure~\ref{fig:exent_q_all_synthetic} shows the results of different variants of \quantmodel and \model across the different task complexities as the relative performance gain over the RoBERTa w/o Exp. baseline.
For ease of visualization, we have averaged the results across binary and multiclass classification tasks in the figure. Post-averaging, we plot sets of four bars corresponding to 
% Each set of four bars in the figure correspond to 
the evaluations of the four models (three \quantmodel variations + \model) on each of the 24 task complexities resulting from negations, conjunctions, and quantifiers. 
% \ssr{This is very unclear -- I see six panels in Fig 3, each having 4 groups -- I don't see 48 task complexities, you need to make this clearer to the reader}

Overall, we find that explicit modeling of quantifier semantics help learn better zero-shot classifiers. In particular, \quantmodel expectedly performs much better than previous approaches on tasks with quantified explanations. Further, while \model is weaker than RoBERTa w/o exp. baseline on certain task complexities, the \quantmodel variants outperform or match the baselines on almost all task complexities.
% \quantmodel (atleast one of the variants) outperforms prior work (\model) in all the synthetic task complexities.
% \quantmodel outperforms RoBERTa w/o exp. baseline for all synthetic task complexitis
% \ssr{What are the task types? Where do they appear in the figure? The description of this figure needs to be significantly expanded} containing quantifiers. 
Expectedly, the generalization ability of models decrease with the increasing complexity of explanations due to changes in the structure of explanations or the presence of negations. 
% \ssr{This is all very hard to follow: make it explicit what are task types and more complex/less complex tasks (by describing it here, but also showing/mentioning it in the figure/caption: that within each panel the complexity is increasing from left to right. ).}

\quantmodel (ordinal) performs the best across the majority of the synthetic task complexities in \benchmark with a significant 5.2\% absolute improvement across all tasks complexities over \model ($p < 0.001$, paired t-test). Further, \quantmodel (predefined init) performs comparably with \quantmodel (ordinal) in many cases (5.0\% vs 5.2\% improvement over \model
but struggles in tasks where explanations have negations in both clauses and labels. 
% \ssr{include some numbers in the text to back your assertions up. Also, have we done any statistical testing here?}
% Out of the three variants, the \quantmodel which learns quantifier semantics from random initialization, \quantmodel (scratch), performs worst, which highlights the challenge of jointly learning quantifier semantics and a classifier only from labels. Even so, we observe that \quantmodel (scratch) outperforms \model by 2\% points on average across all task types. 
% \ssr{Is QuexEnt random  > Exent?}
\quantmodel (random init) performs worst out of the three variants, highlighting the challenge of jointly learning quantifier semantics and a classifier only from labels. Nevertheless, \quantmodel (random init) outperforms \model significantly by 2\% points (absolute; $p < 0.005$, paired t-test) on average across all task complexities.

Analyzing the learned quantifier estimates (Figure \ref{fig:quantifier_weights}), for \quantmodel (predefined init) we observe the final learned probability values to be close to the initialization values. %We hypothesize that since the datasets were also created with the same probability values for quantifiers in \benchmarksyn, the model did not need to alter the estimates too much. 
% Instead, it could focus only on learning the classifier. 
On the other hand, we note that \quantmodel (ordinal) learns three clusters of quantifier probabilities that match with our intuition of high-strength (probability above 0.95), intermediate strength (probability around 0.7), and low-strength quantifiers (probability close to 0). Even though \quantmodel (ordinal) makes little difference between quantifiers within a cluster, we observe that weak supervision in the form of ordinal ranks is sufficient to develop models competent with, even surpassing,  \quantmodel (predefined) that uses predefined initialization.
Finally, we observe \quantmodel (random init) struggles to learn any interpretable ranking for quantifiers. On further analysis, we identify that \quantmodel (random init) is able to learn the quantifier semantics reasonably well for simple binary tasks. However, it struggles to learn reasonable quantifier semantics in the presence of negations, conjunctions, and disjunctions. 
% \ssriv{What would it take to include a minimal model that considers quantifiers as distributions rather than point values? Is there a simple way for this that just extends your current code? How are you parameterizing $p_{\textrm{quant}}$ -- can you just replace the parameter with a beta-distribution, and learn the parameters of the specific beta distribution for each quantifier? i.e. in eqns (1) and (2), replace $p_{quant}$ with $p_{quant} \times P($p_{quant}$ | \textrm{Beta}_{quantifier} )$ }

\subsection{Aggregating explanations with attention}
\label{sec:attention_quexent}
%\sayan{try to increase content here if possible}

To mimic human learning, models need to identify salient explanations that can be potentially useful in classifying an input. As previously mentioned, in the absence of labeled data for a task, previous work on learning from explanations does not differentiate between multiple explanations in terms of their salience and utility for classifying an example. For example, 
%For example, out of the two explanations in Figure~\ref{fig:intro_figure}, the first explanation is more salient as it provides a stronger supervision to classify the input as `poisonous'.
\model averages the class logits from multiple explanations for making predictions, implicitly 
%However, using mean to aggregate class logits resulting from different explanations (as done in prior work, \model for aggregation; \secref{sec:old_model}) considers 
considering all explanations equally salient for classifying an example. %\ssr{We can't talk about salience wrt a specific example, unless the attention mechanism take the example as input as well}%classifying the input. 
% \ssr{Unclear why you are talking about the mean here, unless you  specify that this is what previous work has done, or this would be the obvious alternative. }.
In order to model the varying importance of each explanation towards deciding the class label, we use attention for the aggregation step (see~\secref{sec:old_model}).
We obtain the attention weights by using a feed-forward network over the \texttt{[CLS]} representations obtained from the intermediate NLI model. 
% \rrm{So you look at the attenion from CLS to each explanation?} 
The attention weights are then normalized using softmax. The final aggregated class logits for the label $l$ is $\sum_{j=1}^m{a_jz_j^l}$, where $a_j$ is the attention weight for each explanation $e_j$, and $z_j^l$ denotes the logit for label $l$ using $e_j$. 
The aggregated class logits are converted to probabilities using softmax, and the model is trained using cross-entropy loss.
\begin{figure}[!h]
    \centering
\minipage{0.22\textwidth}
  \includegraphics[scale=0.55]{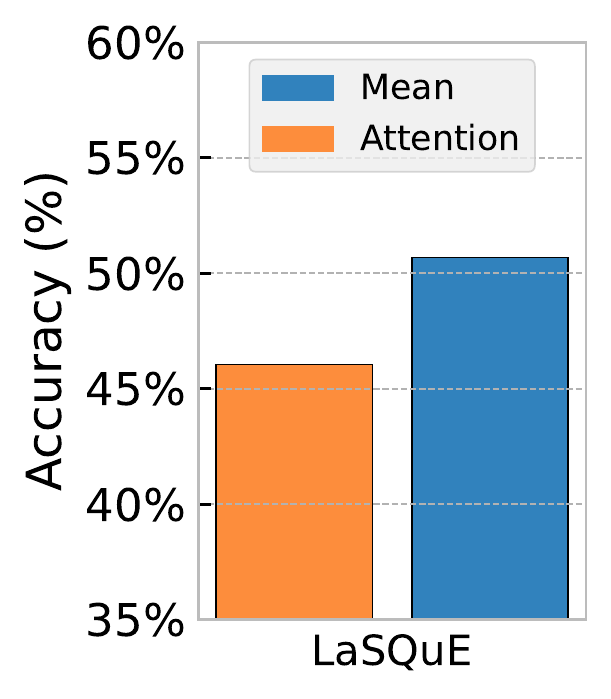}
  \centering
    \begingroup\renewcommand{\caption}[1]{(a)}%
  \caption{}
  \endgroup
\endminipage\hfill\hspace{-2em}
\minipage{0.23\textwidth}%
  \includegraphics[scale=0.55]{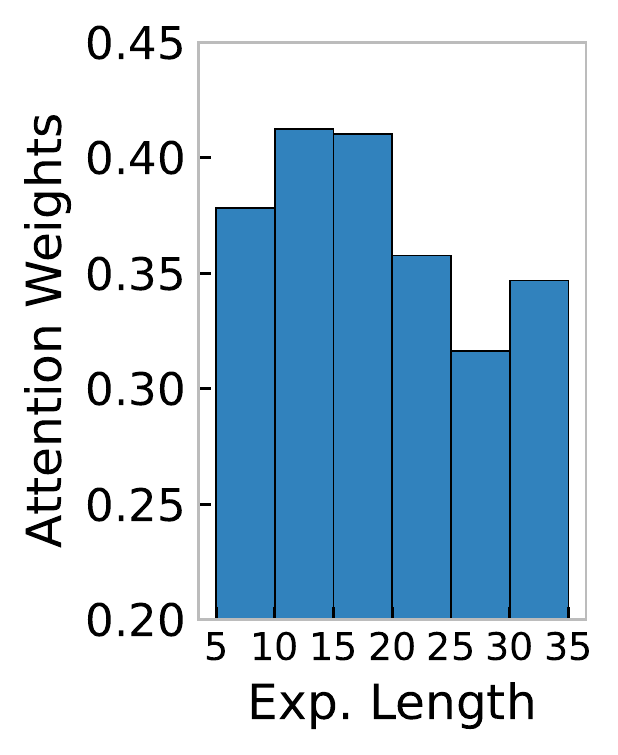}
\centering
  \begingroup\renewcommand{\caption}[1]{(b)}%
  \caption{}
  \endgroup
\endminipage
    \caption{(a) Generalization accuracy on \benchmarksyn ablating the use of attention to combine results from multiple explanations in \quantmodel. (b) Mean attention scores of explanations from \quantmodel vs explanation length (\# of tokens).
    % \rrm{still ordinal} \ssr{Why do we show random init (seems counterintuitive coming from prev section)? Are the others much lower? If we haven't tried those, perhaps not mention the variant here at all (since this subsection is orthogonal to those) }
    }
    % \vspace{-0.4cm}
    \label{fig:attention_plots}
\end{figure}
\begin{figure*}[!ht]
    \centering
    \includegraphics[width=\textwidth]{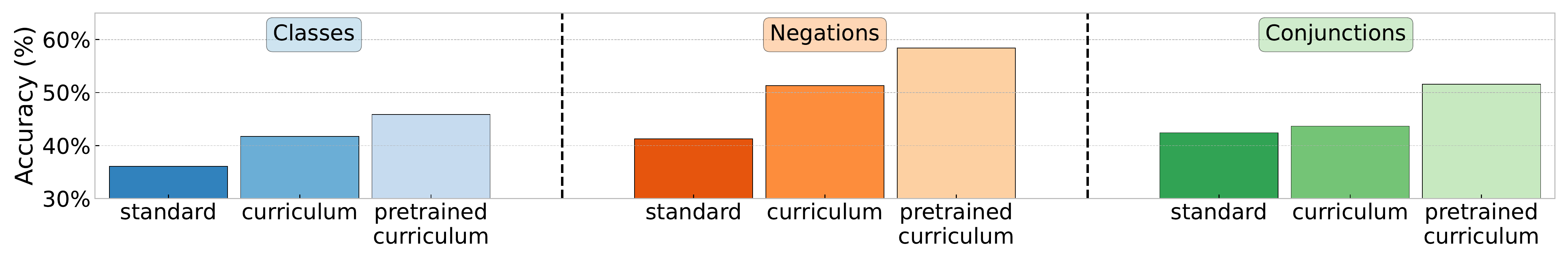}
    \caption{Averaged generalization accuracy on novel classification tasks of the `most difficult' task complexities across three curricula: classes, negations, conjunctions. %Quantifier semantics are pretrained on the simplest synthetic binary tasks (contains no negations/conjunctions/disjunctions). 
    % Curriculum learning, while being effective in all the three curricula, shows maximum gains when handling negations.
    While effective in all three curricula, curriculum learning shows maximum gains when handling negations.
    % \ssr{It is unclear what the three sets of bars are for. You should label each set with the curriculum names (and remove the extra class/neg/conj suffixes from bar names, probably also don't need different color palettes in that case?).}
  }
%   \vspace{-0.2cm}
    \label{fig:exent_q_curriculum_pretrained}
\end{figure*}

\noindent \textbf{Performance on \benchmarksyn:} Figure~\ref{fig:attention_plots}(a) shows the generalization performance on \benchmarksyn for two variants of \quantmodel, one using mean and the other using attention for aggregation. We find that using attention for aggregation across explanations results in significantly better generalization accuracy (50.68\% vs 46.04\% ; $p < 0.1$, paired t-test). While technically simple, we see that this modification allows the model to behave in conceptually sophisticated ways.

\noindent
\textbf{Attention weight analysis: } Figure~\ref{fig:attention_plots}(b) shows a histogram of average attention weights from \quantmodel for different explanation lengths. We find that longer explanations (typically explanations containing nesting of conjunctions and disjunctions) get lower attention weights on average. This seems reasonable and intuitive, since more complex explanations are likely harder for the model to interpret correctly, and hence relying overly on them may be riskier. Further, we find that explanations containing a quantifier receive higher attention on average than explanations without quantifier (0.44 vs 0.35), further highlighting the value of modeling quantifiers in explanations. Explanations containing `definitely' and `frequently' received  higher attention than explanations containing other quantifiers.
Somewhat surprisingly, we found that the average attention weights were comparable for explanations with and without negation. 
% \ssr{Your analysis and results in this section are very interesting findings. You can expand on them a bit further. We should make sure they are not missed -- possibly could also mention something about attention analysis in the intro}
% \ssr{We can include much richer analysis here: e.g., statistics/distribution of attention scores, what types of explanations see high/low attention (both qualitatively, as well as quantitatively -- are attention scores correlated with explanation length, complexity, presence of negation, etc.), etc. Currently, you just say attention works better without any insights into why}
% \sayan{are there any major changes in quantifier semantics of the attention model?}

\subsection{Curriculum learning}
\label{sec:curriculum_learning}

From Figure~\ref{fig:exent_q_all_synthetic}, it is clear that the generalization abilities of models diminish dramatically with the increasing complexity of tasks and explanations. Thus, we next investigate using curriculum learning~\cite{bengio2009curriculum}, which has shown significant successes in learning complex tasks, for learning classifiers from explanations. 
%Drawing motivation from the success of curriculum learning \cite{bengio2009curriculum} in training on an suite of increasingly complex tasks, we explore its utility in learning classifiers from explanations. The decreasing generalization ability of models with increase in explanation complexity (refer Figure~\ref{fig:exent_q_all_synthetic}) further motivates the need of curriculum learning.
 
We define the `complexity' of an explanation under three axes here - (1) the type of classification task (binary vs multiclass) served by the explanation, (2) presence of negations in the explanation, and (3) structure of the explanation (whether the explanation contains conjunction/disjunctions or nested clauses). 
Using curriculum learning we empirically evaluate if training on a classification task with `easier' (less complex) explanations first gives any advantage when learning a task with `harder' (more complex) explanations.
In this work, we explore training \quantmodel under the following curricula:
% \sayan{mention quexent}
\begin{itemize}[noitemsep, topsep=0pt, leftmargin=*]
    \item \uline{Binary $\rightarrow$ multiclass:} We first train classifiers on binary classification tasks and then on multiclass classification tasks. 
    \item \uline{No negations $\rightarrow$ having negations in labels and clauses:} We train on tasks with explanations that contain no negation followed by training on tasks with explanations that have negations in them. Note that negation can appear in the clauses or before a class label in the explanation. 
    
    \item \uline{No conjunctions/disjunctions $\rightarrow$ tasks with nested conjunctions and disjunctions:} 
    We first train on tasks with simple explanations without any conjunctions or disjunctions. Following this, we train on tasks having explanations that contain one conjunction/disjunction and then, on tasks with explanations that contain nested clauses. 
\end{itemize}

\begin{figure*}[!h]
    \centering
    \includegraphics[width=1.0\textwidth]{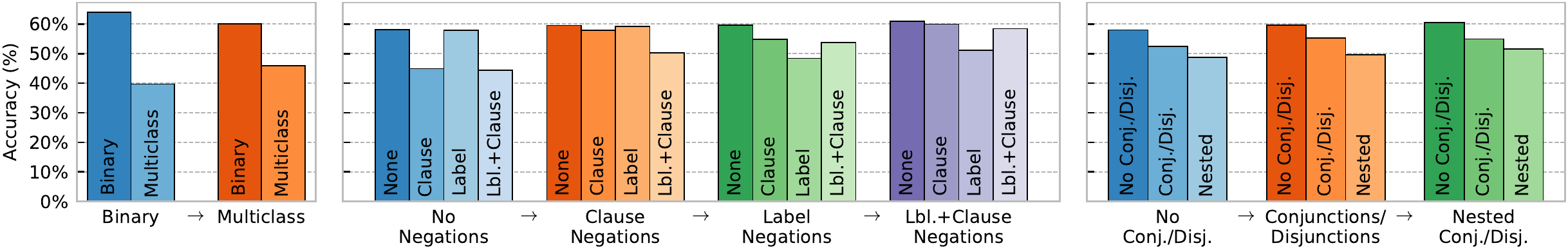}
    \caption{Progression of generalization accuracies on task complexities as we move forward in the curriculum for all three curricula (from left to right: Classes, Negations and Conjunctions curriculum). The text on each bar indicates the evaluation complexity, while the x-axis indicates the complexity that the model has been currently trained on in the curriculum. 
    % \ssr{Text on bars for the last two curricula is too small. You can include abbreviations like in Fig 3 if necessary}
    % Progression of generalization accuracies on task complexities as we move forward in the curriculum. Text on the bar indicates the evaluation set that has been used, while the x-axis indicates the complexity that the model has been currently trained on in the the curriculum. 
    % \ssr{Text too small to read}
  }
  \vspace{-0.2cm}
    \label{fig:exent_q_curriculum_forgetting}
\end{figure*}

Figure~\ref{fig:exent_q_curriculum_pretrained} shows the results of curriculum learning on the synthetic tasks of \benchmark.  We find that \quantmodel trained through curriculum learning (denoted by `curriculum' bars) outperforms \quantmodel trained only on the most challenging task set in the corresponding curriculum (denoted by `standard' bars) on the generalization accuracy of novel hardest tasks in the corresponding curriculum. 
However, we notice that \quantmodel has minimal benefits from training in a curriculum learning fashion on the conjunctions curriculum (shown in green). %in the Figure~\ref{fig:exent_q_curriculum_pretrained}).
% However, the `standard' variants see strictly less training data since they are trained only on the training set of the hardest complexity in the curriculum. Hence, comparing `standard' and `curriculum' numbers is not a fair comparison. Thus, we also include a `multi-task' approach for comparison, which sees as much training data as our `curriculum' learning approach. The `multi-task' variants train \quantmodel on the combined collection of easy and hard tasks of the corresponding curricula. Comparing with the `multi-task' baseline, we observe that the curriculum learning model underperforms on all curricula.
% We find that the `multi-task' model outperforms the `curriculum' trained models in all the three curricula.

% We hypothesize that the curriculum learning models underperform multi-task models here since it has to jointly learn quantifier semantics and to perform classification for tasks of different complexities 
% \ssr{Doesnt the multi-task method have to jointly learn them too? If there is no easy way to resolve this, I would suggest not including the multi-task method here, but still including the pretrained method justified as something like: we hypothesize that jointly learning quantifiers and classifiers might be challenging, so we test what happens if we learn quantifiers of easy tasks first and freeze}
We hypothesize that jointly learning quantifiers and classifiers might be challenging, so we experiment with another setup where we reduce the learning problem to only modeling task complexities by freezing the quantifier semantics with the semantics learned by \quantmodel on simple synthetic binary tasks. With this modification, we find that curriculum learning is much more effective in all three curricula as seen from the improved average generalization performance (denoted by `pretrained curricula' bars in Figure~\ref{fig:exent_q_curriculum_pretrained}). Notably, we find curriculum learning to be most effective in handling negations obtaining an absolute improvement of 17.11\% on the generalization accuracy. The low gains achieved through curriculum learning for handling structural complexity indicates a need to model the role of conjunctions and disjunctions explicitly.
% rather than relying on the neural models to learn them implicitly. 
We leave this for future work to explore.
% \rrm{Need something on the pre-trained bit}\sayan{done}

% \subsubsection{On catastrophic forgetting}

In Figure \ref{fig:exent_q_curriculum_forgetting}, we show the trajectories of generalization performance as we increase the complexity along three independent axes in the three curricula.
% : (a) binary $\rightarrow$ multiclass, (b) no negations $\rightarrow$ having negations in labels and clauses, and (c) no conjunctions/disjunctions $\rightarrow$ tasks with nested conjunctions and disjunctions. 
Briefly, our results indicate that in learning tasks with more classes, generalization increases on multiclass classification tasks at the expense of a slight performance decrease on the more straightforward binary tasks. 
In the curriculum focused on negations, \quantmodel underperforms on tasks with explanations that have `label negations' after training on the relevant training datasets for that complexity. However, on further analysis, we observe that this trend is more pronounced when `label negations' are paired with multiclass classification tasks. By contrast, \quantmodel improves through training on the relevant training datasets of binary classification tasks with `label negations' in concepts.
Lastly, training progressively on more structurally complex tasks resulting from conjunctions/disjunctions in explanations shows improvements during evaluation across all conjunction types without forgetting how to solve simpler tasks. % 

\begin{figure}[!h]
    \centering
    \includegraphics[scale=0.6]{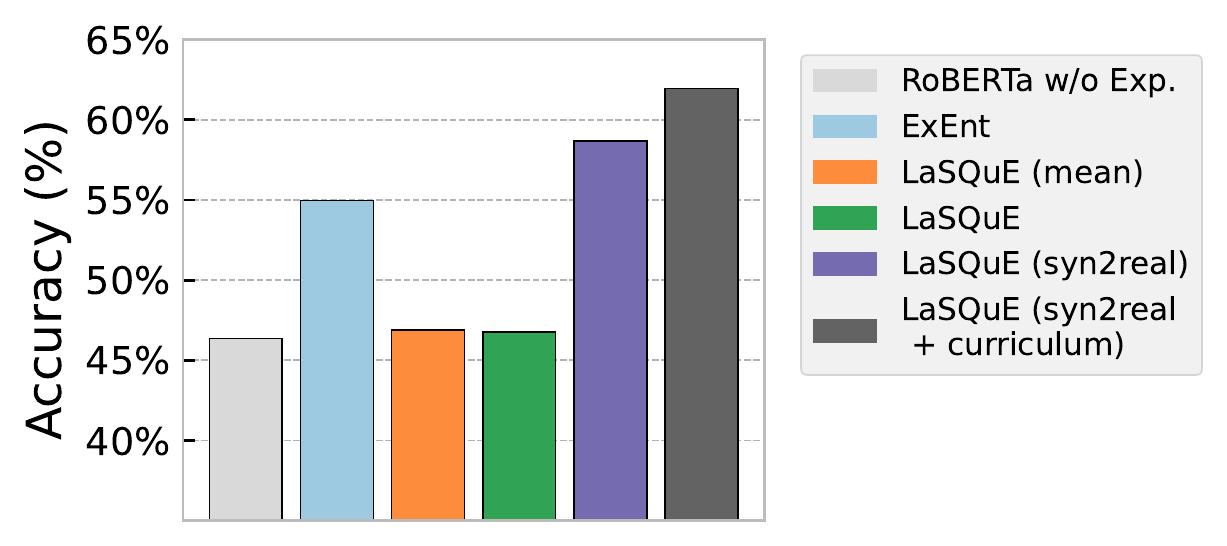}
    \caption{Classification accuracy on novel real-world classification tasks in \benchmarkreal. 
    }
    \label{fig:exent_q_pretraining_for_real}
\end{figure}

\section{Performance on real-world tasks}
\label{sec:real_world_results}
\paragraph{Comparison with ExEnt.} In the previous sections, we established the effectiveness of our proposed strategies on a large number of synthetic tasks from \benchmark. Here, we empirically evaluate \quantmodel on the 36 real-world classification tasks from \benchmark
using the aforementioned strategies. 
Figure~\ref{fig:exent_q_pretraining_for_real} shows the generalization performance of \quantmodel and the baselines on \benchmarkreal. 
% For \benchmarkreal, we consider another baseline (not shown in Figure~\ref{fig:exent_q_pretraining_for_real}) by prompting a large language model, T0-3B \cite{sanh2021multitask} using the explanations. We find that T0-3B~\cite{sanh2021multitask} obtains a zero-shot accuracy of 39.4\%  and 35.4\%  with and without using explanations respectively when prompting on test tasks of \benchmarkreal.

We find that directly trying to train \quantmodel fails to surpass the baselines (even when using attention to aggregate over explanations) as 
% the comparatively low number of explanations in \benchmarkreal hinders the model to jointly learn quantifier semantics and classification from explanations. 
the comparatively low number of explanations in \benchmarkreal hinders the model from learning quantifier semantics and classification from explanations jointly. 
To alleviate this issue, we pre-train on the synthetic tasks and then fine-tune the learned model on the tasks in \benchmarkreal, which we also see as a natural type of curriculum learning. 
We find that pre-training on synthetic tasks (\quantmodel (syn2real)) gives a relative gain of 6.7\% in generalization accuracy over \model. 
Next, we evaluate the utility of curriculum learning on real tasks. We start with a pre-trained \quantmodel on synthetic tasks and then fine-tune it first on binary tasks of \benchmarkreal followed by training on multiclass tasks of \benchmarkreal. We find that curriculum learning results in the best generalizable model (\quantmodel (syn2real + curriculum)) performing significantly better than \model (relative gain of 12.7\%; $p < 0.005$, paired t-test) on \benchmarkreal. 
% \quantmodel (syn2real + curriculum) outperforms naive prompting of T0-3B as well (absolute gain of 22.5\%) showing that our strategies for \quantmodel instill stronger inductive biases into a much smaller model (125M for \quantmodel vs 3B for T0).
% \\
% \textcolor{blue}{
\paragraph{Comparison with Instruction-tuned models.}
% Prompting large language models is understudied for structured inputs. 
Recent works show that `large' language models fine-tuned on multiple classification tasks have an ability for zero-shot classification on new tasks \cite{ouyang2022training, sanh2021multitask}. These models have been primarily trained on unstructured text classification tasks and instructions that define the task rather than providing logic for classification. Given the emergent ability of large language models \cite{wei2022emergent}, we test the performance of such models on \benchmarkreal and compare them with our best \quantmodel model. Specifically, we compare against the publicly available T0-3B \cite{sanh2021multitask}. We also report numbers with and without explanations in the prompt for T0.
%  and FLAN-T5-XXL \cite{chung2022scaling}
% For \benchmarkreal, we consider another baseline by prompting a large language model, T0-3B \cite{sanh2021multitask} using the explanations.
As shown in Table \ref{tab:main}, we find that T0-3B with explanations in the prompt (52.6\%) obtains a zero-shot accuracy lower than the accuracy achieved by our best \quantmodel (61.9\%) on the same set of tasks. This shows that our strategies for \quantmodel instill stronger inductive biases into a much smaller model (125M for \quantmodel vs 3B for T0). 
% However, FLAN-T5-XXL, which is trained on significantly more number of instruction-based tasks than T0 or \quantmodel ($\sim2000$ tasks), outperforms both models by a large margin when prompted with explanations demonstrating the utility of pre-training instruction-tuned models on large number of tasks.
% \ssc{Include comparisons to FLAN and GPT3? Also, instead of prose, including these in prev figures?}
% }
\begin{table}[h!]
\centering
\begin{tabular}{l|c}
\hline
\Thead{Method} & \Thead{Accuracy}\\
\hline
\quantmodel (best) & \textbf{61.9\%}\\
\hline
T0-3B (w/o explanations) & 49.2\%\\
T0-3B (w/ explanations) & 52.6\%\\
\hline
% FLAN-T5-XL (w/o exp) & 55.8\%\\
% FLAN-T5-XL (w/ exp) & \textbf{71.2}\%\\
% \hline
% FLAN-T5-XXL (w/o exp) & 55.9\% % 55.8\%\\
% FLAN-T5-XXL (w/ exp) & \textbf{71.5}\% % 71.25\%\\
% \hline
\end{tabular}
\caption{Zero-shot accuracy of our best \quantmodel and T0 on 16 unseen classification tasks of \benchmarkreal.}
\label{tab:main}
\end{table}

% \ssr{Can also mention more details, such as significance testing, number of tasks out of 36 where gains were seen, tasks with extreme differences, etc.} 
% \ssr{Include and discuss other variants: attention, curriculum}
% explore two approaches -- (1) pre-train quantifier semantics on the simplest binary synthetic tasks and mapping the new quantifiers in real explanation to known quantifiers in synthetic explanations, and (2) pretraining \quantmodel on \benchmarksyn and then fine-tuning the model on \benchmarkreal. Figure~\ref{fig:exent_q_pretraining_for_real} shows the results for these two approaches. We find that pre-training on synthetic tasks (\quantmodel (syn2real) Figure~\ref{fig:exent_q_pretraining_for_real}) leads to the best generalizable \quantmodel on \benchmarkreal.

% \ssr{The following section currently seems a bit out-of-place, I suggest removing it for now, and including a more comprehensive version in the camera-ready version} We analyze the attention scores over explanations from \quantmodel (syn2real), to identify the characteristics of explanations that are considered to be more salient. We find that explanations having conjunction, disjunction or negation have more attention weight on average. Contrary to our observation for synthetic explanations, \quantmodel (syn2real) assigns lower attention score on average to real explanations with quantifiers.
\section{Conclusion}
\label{sec:conclusion}
We have presented effective and generalizable strategies to learn classifiers from language explanations. %Our strategies focus on modeling quantifier semantics, using attention for aggregation across explanations and curriculum learning. 
While our results are promising, our analysis also %model, \quantmodel outperforms prior work showing better generalizing on tasks of the \benchmark benchmark, we
highlights several open challenges in learning from language.
In particular, \quantmodel struggles to learn quantifier semantics without quantifier-specific supervision (in the form of pre-defined initialization or ordinal relations), especially when tasks have complex explanations (due to the presence of negations/conjunctions/disjunctions). Further, our modeling of quantifiers as fixed probability values is restrictive. Future work can also explore explicit modeling of negations, conjunctions and disjunctions for learning from explanations. 
% \ssr{Include Ethics and broader impact statement}

\section{Limitations}

In this work, we introduce \quantmodel, which models and learns the differential semantics of linguistic quantifiers present in natural language explanation to train a classifier guided by these explanations. We evaluate the efficacy of \quantmodel over baselines on the \benchmark benchmark. 

This work assumes that only a single quantifier is present in the explanations.
However, in real-world settings, explanations may contain multiple quantifiers. Modeling the composition of quantifiers can be an interesting direction for future work to make the paradigm of learning from explanations more robust toward fuzzy concepts expressed in real-world explanations.

For our experiments, we assume perfect extraction of quantifiers and limit our analysis to a limited set of quantifiers in this work. %, we extract them through string matching. 
Furthermore, we assume that the effect of quantifiers in a sentence is the same irrespective of the domain of the sentence. For example, consider two sentences \textit{`pungent mushrooms are usually toxic'} and \textit{`people who smoke regularly usually suffer from cancer'}. Here the effect of \emph{`usually'} is not exactly the same for two sentences that are from different domains. However, \quantmodel is not sensitive to the task domain while modeling the semantics of the quantifier. Future work can investigate variations in the semantics of the same quantifier across different domains and also how to incorporate/learn such domain-specific differences (for example, by modeling the semantics of a quantifier as a probability distribution rather than a point value).

\section*{Ethics and Broader Impact}

All our experiments are performed over publicly available datasets, specifically datasets (including language explanations) from \benchmark benchmark~\cite{anon_benchmark}. The datasets do not contain any information that uniquely identifies the crowdworkers involved in data collection. %For details of data collection of \benchmark, we refer the reader to \citet{anon_benchmark}. 
We do not perform any additional annotation or human evaluation in this work. 

Our method, \quantmodel can learn classifiers over structured data using language explanations provided as part of input to the classifier. \quantmodel is built over existing pre-trained language model, RoBERTa~\cite{liu2019roberta}. We do not foresee any risks with our method if the inputs to our model are appropriate for the task. Any measures to counteract erroneous inputs (that may be provided deliberately, potentially exploiting unwanted biases) or curb the biases of pre-trained language models are beyond the scope of this work.

The broader impact of this research in the longer term could increase the accessibility of predictive technologies for ordinary users (non-experts), enabling them to customize AI technologies through natural language interactions.

\bibliography{custom}

\begin{thebibliography}{44}
\expandafter\ifx\csname natexlab\endcsname\relax\def\natexlab#1{#1}\fi

\bibitem[{Andreas et~al.(2018)Andreas, Klein, and
  Levine}]{andreas-etal-2018-learning}
Jacob Andreas, Dan Klein, and Sergey Levine. 2018.
\newblock \href {https://doi.org/10.18653/v1/N18-1197} {Learning with latent
  language}.
\newblock In \emph{Proceedings of the 2018 Conference of the North {A}merican
  Chapter of the Association for Computational Linguistics: Human Language
  Technologies, Volume 1 (Long Papers)}, pages 2166--2179, New Orleans,
  Louisiana. Association for Computational Linguistics.

\bibitem[{Arabshahi et~al.(2020)Arabshahi, Mazaitis, Li, Myers, and
  Mitchell}]{arabshahiconversational}
Forough Arabshahi, Kathryn Mazaitis, Toby Jia-Jun Li, Brad~A Myers, and Tom
  Mitchell. 2020.
\newblock \href
  {https://forougha.github.io/paperPDF/Conversational_Learning.pdf}
  {Conversational learning}.

\bibitem[{Bach et~al.(2013)Bach, Jelinek, Kratzer, and
  Partee}]{bach2013quantification}
Elke Bach, Eloise Jelinek, Angelika Kratzer, and Barbara~BH Partee. 2013.
\newblock \href {https://link.springer.com/book/10.1007/978-94-017-2817-1}
  {\emph{Quantification in natural languages}}, volume~54.
\newblock Springer Science \& Business Media.

\bibitem[{Barwise and Cooper(1981)}]{barwise1981generalized}
Jon Barwise and Robin Cooper. 1981.
\newblock \href {https://www.jstor.org/stable/25001052} {Generalized
  quantifiers and natural language}.
\newblock In \emph{Philosophy, language, and artificial intelligence}, pages
  241--301. Springer.

\bibitem[{Bengio et~al.(2009)Bengio, Louradour, Collobert, and
  Weston}]{bengio2009curriculum}
Yoshua Bengio, J{\'e}r{\^o}me Louradour, Ronan Collobert, and Jason Weston.
  2009.
\newblock \href {https://dl.acm.org/doi/abs/10.1145/1553374.1553380}
  {Curriculum learning}.
\newblock In \emph{Proceedings of the 26th {A}nnual {I}nternational
  {C}onference on {M}achine {L}earning}, pages 41--48.

\bibitem[{Chai et~al.(2020)Chai, Wu, Han, Fei, and
  Li}]{desc_based_classification}
Duo Chai, Wei Wu, Qinghong Han, Wu~Fei, and Jiwei Li. 2020.
\newblock Description based text classification with reinforcement learning.
\newblock In \emph{Proceedings of the 37th International Conference on Machine
  Learning}, ICML'20. JMLR.org.

\bibitem[{Chopra et~al.(2019)Chopra, Tessler, and Goodman}]{Chopra2019TheFC}
Sahil Chopra, Michael~Henry Tessler, and Noah~D. Goodman. 2019.
\newblock \href {https://cogsci.mindmodeling.org/2019/papers/0060/index.html}
  {The first crank of the cultural ratchet: Learning and transmitting concepts
  through language}.
\newblock In \emph{CogSci}.

\bibitem[{Cui et~al.(2022)Cui, Hershcovich, and
  S{\o}gaard}]{cui-etal-2022-generalized-quantifiers}
Ruixiang Cui, Daniel Hershcovich, and Anders S{\o}gaard. 2022.
\newblock \href {https://aclanthology.org/2022.naacl-main.359} {Generalized
  quantifiers as a source of error in multilingual {NLU} benchmarks}.
\newblock In \emph{Proceedings of the 2022 Conference of the North American
  Chapter of the Association for Computational Linguistics: Human Language
  Technologies}, pages 4875--4893, Seattle, United States. Association for
  Computational Linguistics.

\bibitem[{Efrat and Levy(2020)}]{efrat2020turking}
Avia Efrat and Omer Levy. 2020.
\newblock \href {https://arxiv.org/abs/2010.11982} {The turking test: Can
  language models understand instructions?}
\newblock \emph{arXiv preprint arXiv:2010.11982}.

\bibitem[{Guo et~al.(2018)Guo, Huang, Zhang, Zhuang, Dong, Scott, and
  Huang}]{guo2018curriculumnet}
Sheng Guo, Weilin Huang, Haozhi Zhang, Chenfan Zhuang, Dengke Dong, Matthew~R
  Scott, and Dinglong Huang. 2018.
\newblock \href
  {https://www.ecva.net/papers/eccv_2018/papers_ECCV/papers/Sheng_Guo_CurriculumNet_Learning_from_ECCV_2018_paper.pdf}
  {Curriculumnet: Weakly supervised learning from large-scale web images}.
\newblock In \emph{Proceedings of the European Conference on Computer Vision
  (ECCV)}, pages 135--150.

\bibitem[{Hacohen and Weinshall(2019)}]{hacohen2019power}
Guy Hacohen and Daphna Weinshall. 2019.
\newblock \href {https://proceedings.mlr.press/v97/hacohen19a.html} {On the
  power of curriculum learning in training deep networks}.
\newblock In \emph{Proceedings of the 36th International Conference on Machine
  Learning}, volume~97 of \emph{Proceedings of Machine Learning Research},
  pages 2535--2544. PMLR.

\bibitem[{Hancock et~al.(2018)Hancock, Varma, Wang, Bringmann, Liang, and
  R{\'e}}]{hancock-etal-2018-training}
Braden Hancock, Paroma Varma, Stephanie Wang, Martin Bringmann, Percy Liang,
  and Christopher R{\'e}. 2018.
\newblock \href {https://doi.org/10.18653/v1/P18-1175} {Training classifiers
  with natural language explanations}.
\newblock In \emph{Proceedings of the 56th Annual Meeting of the Association
  for Computational Linguistics (Volume 1: Long Papers)}, pages 1884--1895,
  Melbourne, Australia. Association for Computational Linguistics.

\bibitem[{Hanjie et~al.(2022)Hanjie, Deshpande, and
  Narasimhan}]{hanjie2022semantic}
Austin~W Hanjie, Ameet Deshpande, and Karthik Narasimhan. 2022.
\newblock Semantic supervision: Enabling generalization over output spaces.
\newblock \emph{arXiv preprint arXiv:2202.13100}.

\bibitem[{Harris et~al.(2020)Harris, Millman, van~der Walt, Gommers, Virtanen,
  Cournapeau, Wieser, Taylor, Berg, Smith, Kern, Picus, Hoyer, van Kerkwijk,
  Brett, Haldane, del R{\'{i}}o, Wiebe, Peterson, G{\'{e}}rard-Marchant,
  Sheppard, Reddy, Weckesser, Abbasi, Gohlke, and Oliphant}]{harris2020array}
Charles~R. Harris, K.~Jarrod Millman, St{\'{e}}fan~J. van~der Walt, Ralf
  Gommers, Pauli Virtanen, David Cournapeau, Eric Wieser, Julian Taylor,
  Sebastian Berg, Nathaniel~J. Smith, Robert Kern, Matti Picus, Stephan Hoyer,
  Marten~H. van Kerkwijk, Matthew Brett, Allan Haldane, Jaime~Fern{\'{a}}ndez
  del R{\'{i}}o, Mark Wiebe, Pearu Peterson, Pierre G{\'{e}}rard-Marchant,
  Kevin Sheppard, Tyler Reddy, Warren Weckesser, Hameer Abbasi, Christoph
  Gohlke, and Travis~E. Oliphant. 2020.
\newblock \href {https://doi.org/10.1038/s41586-020-2649-2} {Array programming
  with {NumPy}}.
\newblock \emph{Nature}, 585(7825):357--362.

\bibitem[{Jiang et~al.(2018)Jiang, Zhou, Leung, Li, and
  Fei-Fei}]{jiang2018mentornet}
Lu~Jiang, Zhengyuan Zhou, Thomas Leung, Li-Jia Li, and Li~Fei-Fei. 2018.
\newblock \href {https://proceedings.mlr.press/v80/jiang18c.html} {Mentornet:
  Learning data-driven curriculum for very deep neural networks on corrupted
  labels}.
\newblock In \emph{International Conference on Machine Learning}, pages
  2304--2313. PMLR.

\bibitem[{Jones et~al.(2001--)Jones, Oliphant, Peterson
  et~al.}]{jones2001scipy}
Eric Jones, Travis Oliphant, Pearu Peterson, et~al. 2001--.
\newblock \href {http://www.scipy.org/} {{SciPy}: Open source scientific tools
  for {Python}}.

\bibitem[{Kurtzman and MacDonald(1993)}]{kurtzman1993resolution}
Howard~S Kurtzman and Maryellen~C MacDonald. 1993.
\newblock \href
  {https://citeseerx.ist.psu.edu/viewdoc/download?doi=10.1.1.460.4104&rep=rep1&type=pdf}
  {Resolution of quantifier scope ambiguities}.
\newblock \emph{Cognition}, 48(3):243--279.

\bibitem[{Liu et~al.(2019)Liu, Ott, Goyal, Du, Joshi, Chen, Levy, Lewis,
  Zettlemoyer, and Stoyanov}]{liu2019roberta}
Yinhan Liu, Myle Ott, Naman Goyal, Jingfei Du, Mandar Joshi, Danqi Chen, Omer
  Levy, Mike Lewis, Luke Zettlemoyer, and Veselin Stoyanov. 2019.
\newblock \href {https://arxiv.org/abs/1907.11692} {Roberta: A robustly
  optimized bert pretraining approach}.
\newblock \emph{arXiv preprint arXiv:1907.11692}.

\bibitem[{Lobner(1986)}]{lobner1986quantification}
Sebastian Lobner. 1986.
\newblock \href
  {https://www.degruyter.com/document/doi/10.1515/9783112420027-004/html}
  {Quantification as a major module of natural language semantics}.
\newblock In \emph{Studies in discourse representation theory and the theory of
  generalized quantifiers}, pages 53--86. De Gruyter.

\bibitem[{Loshchilov and Hutter(2019)}]{loshchilov2018decoupled}
Ilya Loshchilov and Frank Hutter. 2019.
\newblock \href {https://openreview.net/forum?id=Bkg6RiCqY7} {Decoupled weight
  decay regularization}.
\newblock In \emph{International Conference on Learning Representations}.

\bibitem[{Mao et~al.(2019)Mao, Gan, Kohli, Tenenbaum, and Wu}]{mao2018neuro}
Jiayuan Mao, Chuang Gan, Pushmeet Kohli, Joshua~B. Tenenbaum, and Jiajun Wu.
  2019.
\newblock \href {https://openreview.net/forum?id=rJgMlhRctm} {The
  neuro-symbolic concept learner: Interpreting scenes, words, and sentences
  from natural supervision}.
\newblock In \emph{International Conference on Learning Representations}.

\bibitem[{Mei et~al.(2022)Mei, Mao, Wang, Gan, and Tenenbaum}]{mei2022falcon}
Lingjie Mei, Jiayuan Mao, Ziqi Wang, Chuang Gan, and Joshua~B. Tenenbaum. 2022.
\newblock \href {https://openreview.net/forum?id=htWIlvDcY8} {{FALCON}: Fast
  visual concept learning by integrating images, linguistic descriptions, and
  conceptual relations}.
\newblock In \emph{International Conference on Learning Representations}.

\bibitem[{Menon et~al.(2022)Menon, Ghosh, and Srivastava}]{anon_benchmark}
Rakesh~R Menon, Sayan Ghosh, and Shashank Srivastava. 2022.
\newblock \href {https://doi.org/10.18653/v1/2022.acl-long.451} {{CLUES}: A
  benchmark for learning classifiers using natural language explanations}.
\newblock In \emph{Proceedings of the 60th Annual Meeting of the Association
  for Computational Linguistics (Volume 1: Long Papers)}, pages 6523--6546,
  Dublin, Ireland. Association for Computational Linguistics.

\bibitem[{Mishra et~al.(2022)Mishra, Khashabi, Baral, and
  Hajishirzi}]{Mishra2021CrossTaskGV}
Swaroop Mishra, Daniel Khashabi, Chitta Baral, and Hannaneh Hajishirzi. 2022.
\newblock \href {https://arxiv.org/abs/2104.08773} {Cross-task generalization
  via natural language crowdsourcing instructions}.
\newblock In \emph{Proceedings of the 60th Annual Meeting of the Association
  for Computational Linguistics (to appear)}.

\bibitem[{Moxey and Sanford(1986)}]{quantifier_focus}
Linda~M. Moxey and Anthony~J. Sanford. 1986.
\newblock \href {https://doi.org/10.1093/jos/5.3.189} {{Quantifiers and
  Focus}}.
\newblock \emph{Journal of Semantics}, 5(3):189--206.

\bibitem[{Moxey and Sanford(1993)}]{moxey1993prior}
Linda~M Moxey and Anthony~J Sanford. 1993.
\newblock \href {https://www.tandfonline.com/doi/abs/10.1080/09541449308406515}
  {Prior expectation and the interpretation of natural language quantifiers}.
\newblock \emph{European Journal of Cognitive Psychology}, 5(1):73--91.

\bibitem[{Newport(1990)}]{newport1990maturational}
Elissa~L Newport. 1990.
\newblock \href
  {https://onlinelibrary.wiley.com/doi/abs/10.1207/s15516709cog1401_2}
  {Maturational constraints on language learning}.
\newblock \emph{Cognitive science}, 14(1):11--28.

\bibitem[{Obeidat et~al.(2019)Obeidat, Fern, Shahbazi, and
  Tadepalli}]{obeidat-etal-2019-description}
Rasha Obeidat, Xiaoli Fern, Hamed Shahbazi, and Prasad Tadepalli. 2019.
\newblock \href {https://doi.org/10.18653/v1/N19-1087} {Description-based
  zero-shot fine-grained entity typing}.
\newblock In \emph{Proceedings of the 2019 Conference of the North {A}merican
  Chapter of the Association for Computational Linguistics: Human Language
  Technologies, Volume 1 (Long and Short Papers)}, pages 807--814, Minneapolis,
  Minnesota. Association for Computational Linguistics.

\bibitem[{Ouyang et~al.(2022)Ouyang, Wu, Jiang, Almeida, Wainwright, Mishkin,
  Zhang, Agarwal, Slama, Ray et~al.}]{ouyang2022training}
Long Ouyang, Jeff Wu, Xu~Jiang, Diogo Almeida, Carroll~L Wainwright, Pamela
  Mishkin, Chong Zhang, Sandhini Agarwal, Katarina Slama, Alex Ray, et~al.
  2022.
\newblock Training language models to follow instructions with human feedback.
\newblock \emph{arXiv preprint arXiv:2203.02155}.

\bibitem[{Paszke et~al.(2019)Paszke, Gross, Massa, Lerer, Bradbury, Chanan,
  Killeen, Lin, Gimelshein, Antiga, Desmaison, Kopf, Yang, DeVito, Raison,
  Tejani, Chilamkurthy, Steiner, Fang, Bai, and Chintala}]{NEURIPS2019_9015}
Adam Paszke, Sam Gross, Francisco Massa, Adam Lerer, James Bradbury, Gregory
  Chanan, Trevor Killeen, Zeming Lin, Natalia Gimelshein, Luca Antiga, Alban
  Desmaison, Andreas Kopf, Edward Yang, Zachary DeVito, Martin Raison, Alykhan
  Tejani, Sasank Chilamkurthy, Benoit Steiner, Lu~Fang, Junjie Bai, and Soumith
  Chintala. 2019.
\newblock \href
  {http://papers.neurips.cc/paper/9015-pytorch-an-imperative-style-high-performance-deep-learning-library.pdf}
  {Pytorch: An imperative style, high-performance deep learning library}.
\newblock In H.~Wallach, H.~Larochelle, A.~Beygelzimer, F.~d\textquotesingle
  Alch\'{e}-Buc, E.~Fox, and R.~Garnett, editors, \emph{Advances in Neural
  Information Processing Systems 32}, pages 8024--8035. Curran Associates, Inc.

\bibitem[{Pavlakos et~al.(2018)Pavlakos, Zhou, and
  Daniilidis}]{pavlakos2018ordinal}
Georgios Pavlakos, Xiaowei Zhou, and Kostas Daniilidis. 2018.
\newblock \href
  {https://openaccess.thecvf.com/content_cvpr_2018/html/Pavlakos_Ordinal_Depth_Supervision_CVPR_2018_paper.html}
  {Ordinal depth supervision for 3d human pose estimation}.
\newblock In \emph{Proceedings of the IEEE Conference on Computer Vision and
  Pattern Recognition}, pages 7307--7316.

\bibitem[{Platanios et~al.(2019)Platanios, Stretcu, Neubig, Poczos, and
  Mitchell}]{platanios2019competence}
Emmanouil~Antonios Platanios, Otilia Stretcu, Graham Neubig, Barnabas Poczos,
  and Tom Mitchell. 2019.
\newblock \href {https://doi.org/10.18653/v1/N19-1119} {Competence-based
  curriculum learning for neural machine translation}.
\newblock In \emph{Proceedings of the 2019 Conference of the North {A}merican
  Chapter of the Association for Computational Linguistics: Human Language
  Technologies, Volume 1 (Long and Short Papers)}, pages 1162--1172,
  Minneapolis, Minnesota. Association for Computational Linguistics.

\bibitem[{Sanh et~al.(2022)Sanh, Webson, Raffel, Bach, Sutawika, Alyafeai,
  Chaffin, Stiegler, Raja, Dey, Bari, Xu, Thakker, Sharma, Szczechla, Kim,
  Chhablani, Nayak, Datta, Chang, Jiang, Wang, Manica, Shen, Yong, Pandey,
  Bawden, Wang, Neeraj, Rozen, Sharma, Santilli, Fevry, Fries, Teehan, Scao,
  Biderman, Gao, Wolf, and Rush}]{sanh2021multitask}
Victor Sanh, Albert Webson, Colin Raffel, Stephen Bach, Lintang Sutawika, Zaid
  Alyafeai, Antoine Chaffin, Arnaud Stiegler, Arun Raja, Manan Dey, M~Saiful
  Bari, Canwen Xu, Urmish Thakker, Shanya~Sharma Sharma, Eliza Szczechla,
  Taewoon Kim, Gunjan Chhablani, Nihal Nayak, Debajyoti Datta, Jonathan Chang,
  Mike Tian-Jian Jiang, Han Wang, Matteo Manica, Sheng Shen, Zheng~Xin Yong,
  Harshit Pandey, Rachel Bawden, Thomas Wang, Trishala Neeraj, Jos Rozen,
  Abheesht Sharma, Andrea Santilli, Thibault Fevry, Jason~Alan Fries, Ryan
  Teehan, Teven~Le Scao, Stella Biderman, Leo Gao, Thomas Wolf, and Alexander~M
  Rush. 2022.
\newblock \href {https://openreview.net/forum?id=9Vrb9D0WI4} {Multitask
  prompted training enables zero-shot task generalization}.
\newblock In \emph{International Conference on Learning Representations}.

\bibitem[{Solt(2009)}]{Solt2009THESO}
Stephanie Solt. 2009.
\newblock \href
  {https://citeseerx.ist.psu.edu/viewdoc/download?doi=10.1.1.473.289&rep=rep1&type=pdf}
  {The semantics of adjectives of quantity}.

\bibitem[{Srivastava et~al.(2017)Srivastava, Labutov, and
  Mitchell}]{srivastava2017joint}
Shashank Srivastava, Igor Labutov, and Tom Mitchell. 2017.
\newblock \href {https://doi.org/10.18653/v1/D17-1161} {Joint concept learning
  and semantic parsing from natural language explanations}.
\newblock In \emph{Proceedings of the 2017 Conference on Empirical Methods in
  Natural Language Processing}, pages 1527--1536, Copenhagen, Denmark.
  Association for Computational Linguistics.

\bibitem[{Srivastava et~al.(2018)Srivastava, Labutov, and
  Mitchell}]{srivastava-etal-2018-zero}
Shashank Srivastava, Igor Labutov, and Tom Mitchell. 2018.
\newblock \href {https://doi.org/10.18653/v1/P18-1029} {Zero-shot learning of
  classifiers from natural language quantification}.
\newblock In \emph{Proceedings of the 56th Annual Meeting of the Association
  for Computational Linguistics (Volume 1: Long Papers)}, pages 306--316,
  Melbourne, Australia. Association for Computational Linguistics.

\bibitem[{Steinert-Threlkeld(2021)}]{steinert2021quantifiers}
Shane Steinert-Threlkeld. 2021.
\newblock \href {https://www.mdpi.com/1099-4300/23/10/1335} {Quantifiers in
  natural language: Efficient communication and degrees of semantic
  universals}.
\newblock \emph{Entropy}, 23(10):1335.

\bibitem[{Tay et~al.(2019)Tay, Wang, Luu, Fu, Phan, Yuan, Rao, Hui, and
  Zhang}]{tay2019simple}
Yi~Tay, Shuohang Wang, Anh~Tuan Luu, Jie Fu, Minh~C. Phan, Xingdi Yuan, Jinfeng
  Rao, Siu~Cheung Hui, and Aston Zhang. 2019.
\newblock \href {https://doi.org/10.18653/v1/P19-1486} {Simple and effective
  curriculum pointer-generator networks for reading comprehension over long
  narratives}.
\newblock In \emph{Proceedings of the 57th Annual Meeting of the Association
  for Computational Linguistics}, pages 4922--4931, Florence, Italy.
  Association for Computational Linguistics.

\bibitem[{Wang et~al.(2018)Wang, Singh, Michael, Hill, Levy, and
  Bowman}]{wang-etal-2018-glue}
Alex Wang, Amanpreet Singh, Julian Michael, Felix Hill, Omer Levy, and Samuel
  Bowman. 2018.
\newblock \href {https://doi.org/10.18653/v1/W18-5446} {{GLUE}: A multi-task
  benchmark and analysis platform for natural language understanding}.
\newblock In \emph{Proceedings of the 2018 {EMNLP} Workshop {B}lackbox{NLP}:
  Analyzing and Interpreting Neural Networks for {NLP}}, pages 353--355,
  Brussels, Belgium. Association for Computational Linguistics.

\bibitem[{Wei et~al.(2022)Wei, Tay, Bommasani, Raffel, Zoph, Borgeaud,
  Yogatama, Bosma, Zhou, Metzler, Chi, Hashimoto, Vinyals, Liang, Dean, and
  Fedus}]{wei2022emergent}
Jason Wei, Yi~Tay, Rishi Bommasani, Colin Raffel, Barret Zoph, Sebastian
  Borgeaud, Dani Yogatama, Maarten Bosma, Denny Zhou, Donald Metzler, Ed~H.
  Chi, Tatsunori Hashimoto, Oriol Vinyals, Percy Liang, Jeff Dean, and William
  Fedus. 2022.
\newblock \href {https://openreview.net/forum?id=yzkSU5zdwD} {Emergent
  abilities of large language models}.
\newblock \emph{Transactions on Machine Learning Research}.
\newblock Survey Certification.

\bibitem[{Weller et~al.(2020)Weller, Lourie, Gardner, and
  Peters}]{weller-etal-2020-learning}
Orion Weller, Nicholas Lourie, Matt Gardner, and Matthew~E. Peters. 2020.
\newblock \href {https://doi.org/10.18653/v1/2020.emnlp-main.105} {Learning
  from task descriptions}.
\newblock In \emph{Proceedings of the 2020 Conference on Empirical Methods in
  Natural Language Processing (EMNLP)}, pages 1361--1375, Online. Association
  for Computational Linguistics.

\bibitem[{Wolf et~al.(2020)Wolf, Debut, Sanh, Chaumond, Delangue, Moi, Cistac,
  Rault, Louf, Funtowicz, Davison, Shleifer, von Platen, Ma, Jernite, Plu, Xu,
  Le~Scao, Gugger, Drame, Lhoest, and Rush}]{wolf-etal-2020-transformers}
Thomas Wolf, Lysandre Debut, Victor Sanh, Julien Chaumond, Clement Delangue,
  Anthony Moi, Pierric Cistac, Tim Rault, Remi Louf, Morgan Funtowicz, Joe
  Davison, Sam Shleifer, Patrick von Platen, Clara Ma, Yacine Jernite, Julien
  Plu, Canwen Xu, Teven Le~Scao, Sylvain Gugger, Mariama Drame, Quentin Lhoest,
  and Alexander Rush. 2020.
\newblock \href {https://doi.org/10.18653/v1/2020.emnlp-demos.6} {Transformers:
  State-of-the-art natural language processing}.
\newblock In \emph{Proceedings of the 2020 Conference on Empirical Methods in
  Natural Language Processing: System Demonstrations}, pages 38--45, Online.
  Association for Computational Linguistics.

\bibitem[{Xu et~al.(2020)Xu, Zhang, Mao, Wang, Xie, and
  Zhang}]{xu-etal-2020-curriculum}
Benfeng Xu, Licheng Zhang, Zhendong Mao, Quan Wang, Hongtao Xie, and Yongdong
  Zhang. 2020.
\newblock \href {https://doi.org/10.18653/v1/2020.acl-main.542} {Curriculum
  learning for natural language understanding}.
\newblock In \emph{Proceedings of the 58th Annual Meeting of the Association
  for Computational Linguistics}, pages 6095--6104, Online. Association for
  Computational Linguistics.

\bibitem[{Yildirim et~al.(2013)Yildirim, Degen, Tanenhaus, and
  Jaeger}]{yildirim2013linguistic}
Ilker Yildirim, Judith Degen, Michael Tanenhaus, and Florian Jaeger. 2013.
\newblock \href {https://escholarship.org/uc/item/7h5003md} {Linguistic
  variability and adaptation in quantifier meanings}.
\newblock In \emph{Proceedings of the Annual Meeting of the Cognitive Science
  Society}, volume~35.

\end{thebibliography}
\bibliographystyle{acl_natbib}

\clearpage

\appendix
\appendix

\section{Details on \benchmark} \label{sec:appendix}

\benchmark \cite{anon_benchmark} is a recently proposed benchmark of classification tasks paired with natural language explanations. The benchmark consists of 36 real-world classification tasks (\benchmarkreal) as well 144 synthetic classification tasks (\benchmarksyn). The explanations for the real-world tasks are obtained through crowd-sourcing while they are programmatically generated for \benchmarksyn. In this work, we follow the train and test splits for \benchmarkreal from \citet{anon_benchmark}. Additionally, we train on 70\% of the labeled examples of the seen tasks and perform zero-shot generalization test over the 20\% examples of each task in \benchmarkreal. For the extremely small tasks, we use the entire set of examples for zero-shot testing. The seen-unseen task splits for \benchmarkreal and \benchmarksyn that we use for experiments in this paper is the same as that in \citet{anon_benchmark}.

\subsection{List of quantifiers} \label{sec:tab_quantifiers}
The full list of quantifiers along with their associated probability values are shown in Table~\ref{tab:quantifier}.
\begin{table}[!h]
\small
    \centering
    \begin{tabular}{l|c}
    \toprule
    \Thead{Quantifiers} & \Thead{Probability}\\
    \midrule
    "always", "certainly", "definitely" & 0.95 \\
    \makecell{"usually", "normally", "generally", \\ "likely", "typically"} & 0.70  \\
    "often" & 0.50 \\
     "sometimes", "frequently", & 0.30\\
     "occasionally" & 0.20 \\
    "rarely", "seldom" & 0.10\\
     "never" & 0.05\\
     \bottomrule
    \end{tabular}
    \caption{Probability values used for quantifiers in \benchmark. These values are based on \citet{srivastava-etal-2018-zero}.}
    \label{tab:quantifier}
\end{table}

\section{Training details} \label{sec:model_appendix}
In this section we provide details about implementation such as hyperparameter details, and details about hardware and software used along with an estimate of time taken to train the models.
We provide our implementation code in supplementary material and it will be made public upon first publication.

\subsection{Hyper-parameter settings}
For all the transformer-based models we use the implementation of HuggingFace library \cite{wolf-etal-2020-transformers}. All the model based hyper-parameters are thus kept default to the settings in the HuggingFace library. We use the publicly available checkpoints to initialize the pre-trained models.
For RoBERTa based baselines we use `roberta-base' checkpoint available on HuggingFace.
For our intermediate entailment model in \model, we fine-tune a pretrained checkpoint of RoBERTa trained on MNLI corpus (`textattack/roberta-base-MNLI' from HuggingFace). 

When training on \benchmarksyn, we use a maximum of 64 tokens for our baseline RoBERTa w/o Exp. and \model.

We used the AdamW \cite{loshchilov2018decoupled} optimizer commonly used to fine-tune pre-trained Masked Language Models (MLM) models. 
For fine-tuning the pre-trained models on our benchmark tasks, we experimented with a learning rate of $1e-5$. In order to learn the quantifier probabilities, we search for the correct learning rate to use in $\{1e-3, 2e-3, 5e-3, 9e-3, 1e-2, 2e-2, 3e-2\}$ and use $1e-2$ for our reported experiments based on the best validation accuracy obtained while training and testing on the binary classification datasets with no negation and conjunction complexities in explanations/concepts.
Batch sizes was kept as 2 with gradient accumulation factor of 8. The random seed for all experiments was 42. 
We train all the models for 20 epochs. Each epoch comprises of 100 batches, and in each batch the models look at one of the tasks (in a sequential order) in the seen split. In the curriculum learning experiments, we run the model on each task type for 20 epochs and select the best model during a particular step of the curriculum based on the validation scores of the seen tasks. Finally, the chosen best checkpoint is used to initialize the model for the next step of the curriculum.

\subsection{Hardware and software specifications}
All the models are coded using Pytorch 1.4.0\footnote{\url{https://pytorch.org/}} \cite{NEURIPS2019_9015} and related libraries like numpy \cite{harris2020array}, scipy \cite{jones2001scipy} etc. We run all experiments on a Tesla V100-SXM2 GPU of size 16GB, 250 GB RAM and 40 CPU cores. 

\subsection{Training times}
\begin{itemize}[noitemsep, topsep=0pt, leftmargin=*]
    \item \uline{Training on \benchmarkreal}: The baseline RoBERTa w/o Exp model typically takes 3 seconds on average for training on 1 batch of examples. \model and \quantmodel (all variants) also take comparable amount of time to train on 1 batch. In 1 batch, the models go through 16 examples from the tasks in seen split. 
    \item \uline{Training on \benchmarksyn}: All the models take comparatively much lesser time for training on our synthetic tasks owing to lesser number of explanations on average for a task. For training on 1 batch, all models took 1 seconds or less to train on 1 batch of examples from \benchmarksyn.
    \item \uline{Training for curriculum learning}: The run time of a curriculum learning episode depends on the number of tasks in an episode. In Figure \ref{fig:exent_q_curriculum_pretrained}, the binary-multiclass curriculum takes 2 hours to train, while negations take 4 hours, and conjunctions take 3 hours. The same time frame applies for the results in Figure \ref{fig:exent_q_curriculum_forgetting}.
\end{itemize}

\end{document}